\documentclass[11pt]{article}

\usepackage[final]{acl}

\usepackage{times}
\usepackage{latexsym}

\usepackage[T1]{fontenc}

\usepackage[utf8]{inputenc}
\usepackage{bbm}

\usepackage{microtype}

\usepackage{inconsolata}

\usepackage{graphicx}

\usepackage{hyperref}
\usepackage{url}
\usepackage{algorithm}
\usepackage{algorithmic}
\usepackage{subfiles}
\usepackage{booktabs}       
\usepackage{amsmath,mathtools}
\usepackage{amssymb}
\usepackage{amsfonts}

\usepackage{multirow}
\usepackage{multicol}
\usepackage{xcolor}
\usepackage{adjustbox}
\usepackage{amsthm}
\theoremstyle{definition}

\theoremstyle{remark}

\usepackage{pifont}
\usepackage{caption}
\usepackage{subcaption}
\usepackage{wrapfig}
\usepackage{simplebnf}
\usepackage{tcolorbox}
\usepackage{tikz}
\usepackage{tikz-qtree}
\usepackage{dashrule}
\tcbuselibrary{listings,breakable}
\tcbset{listing engine=listings,colframe=black,colback=white,size=small}
\usepackage{enumitem}

\usepackage{booktabs,multirow,tabularx,makecell,array}
\newcolumntype{Y}{>{\raggedright\arraybackslash}X} 

\usepackage[table]{xcolor}

\usepackage{graphicx}
\usepackage{svg} 

\definecolor{lightgreen}{RGB}{220,245,220}
\definecolor{lightred}{RGB}{255,220,220}

\newcommand{\cmark}{\cellcolor{lightgreen}\checkmark}
\newcommand{\xmark}{\cellcolor{lightred}\texttimes}

\title{Argument Reconstruction as Supervision for Critical Thinking in LLMs}

\author{Hyun Ryu$^{1,2}$\thanks{Equal contribution.},
Gyouk Chu$^{2}$\footnotemark[1],
Gregor Betz$^{3}$,
Eunho Yang$^{2}$,
Carolyn Ros\'e$^{1}$\thanks{Equal advising.},
Sean Welleck$^{1}$\footnotemark[2]\\
$^{1}$Carnegie Mellon University Language Technologies Institute~~~$^{2}$KAIST AI~~~$^{3}$KIT \\
\texttt{\{hyunr,cp3a,swelleck\}@cs.cmu.edu},~~~\texttt{\{ryuhyun1905,kyouwook,eunhoy\}@kaist.ac.kr},\\
\texttt{gregor.betz@kit.edu}
}

\begin{document}
\maketitle
\begin{abstract}
\vspace{-3mm}
To think critically about arguments, human learners are trained to identify, reconstruct, and evaluate arguments. Argument reconstruction is especially important because it makes an argument’s underlying inferences explicit.
However, it remains unclear whether LLMs can similarly enhance their critical thinking ability by learning to reconstruct arguments.
To address this question, we introduce a holistic framework with three contributions. We (1) propose an engine that automatically reconstructs arbitrary arguments (GAAR), (2) synthesize a new high-quality argument reconstruction dataset (Arguinas) using the GAAR engine, and (3) investigate whether learning argument reconstruction benefits downstream critical thinking tasks.
Our experimental results show that, across seven critical thinking tasks, models trained to learn argument reconstruction outperform models that do not, with the largest performance gains observed when training on the proposed Arguinas dataset.
The source code and data will be publicly available.
\vspace{-5mm}
\end{abstract}

\section{Introduction}
\label{sec:introduction}

\begin{figure}[t]
  \vspace{-2mm}
  \includegraphics[width=\columnwidth]{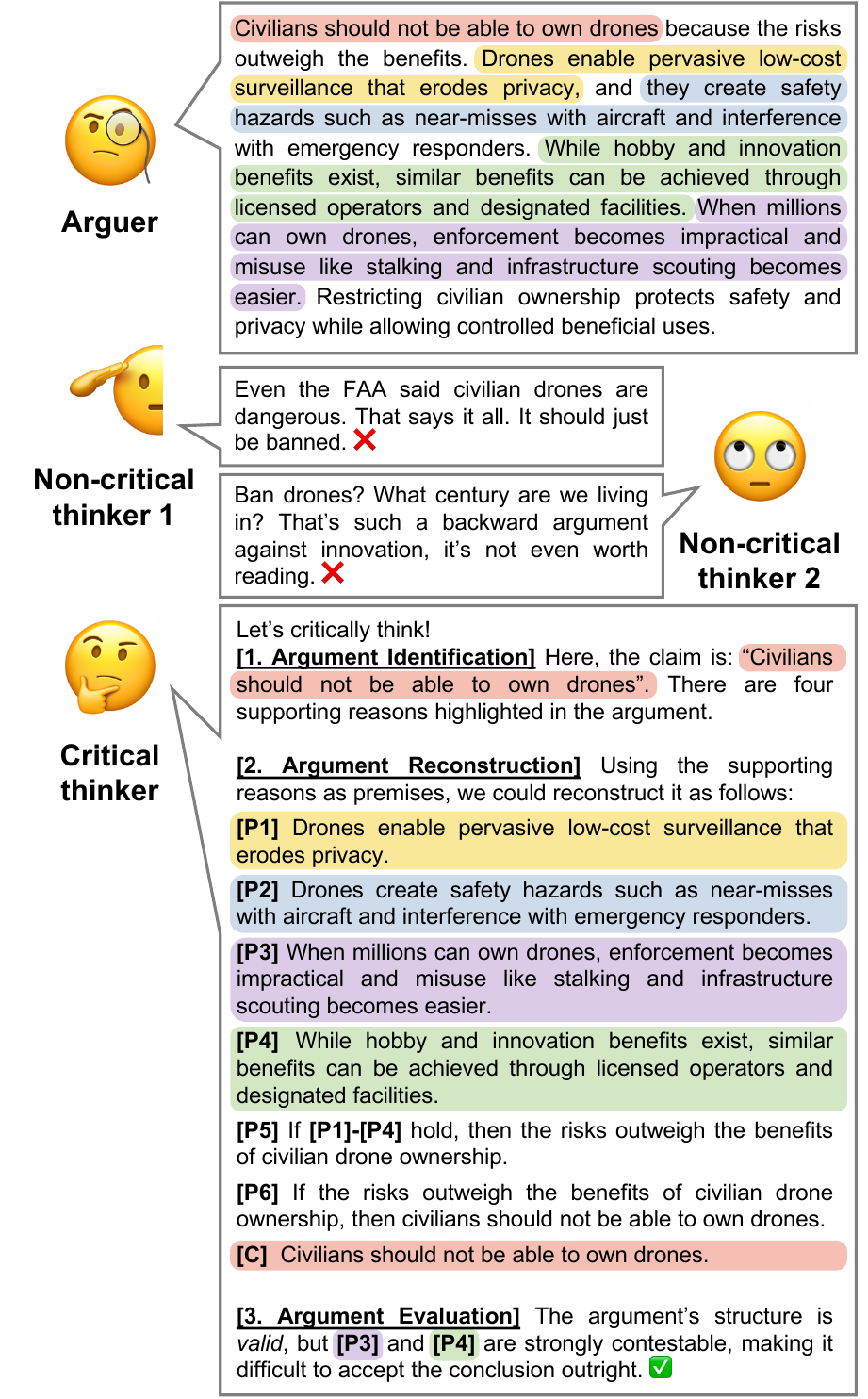}
  \vspace{-5mm}
  \captionof{figure}{Thinking critically about an argument through argument reconstruction.}
  \label{fig:aquinas}
   \vspace{-7mm}
\end{figure}

\textit{Critical thinking} is defined as the ability to think clearly and rationally about what to do or what to believe~\citep{ct-def-1, ct-def-2, ct-def-3}.
Consistent with common treatments for human learners, we adopt the same three-stage operationalization of critical thinking: argument identification, argument reconstruction, and argument evaluation~\citep{ct-def-1, ct-def-2, ct-def-3, ct-3steps-1, ct-3steps-2, ct-3steps-3, ct-3steps-4, ct-3steps-5}.
Specifically, \textit{argument reconstruction} is the process of extracting all explicit and implicit premises and structuring their logical connections, resulting in a \textit{valid} premise--conclusion structure\footnote{A reconstruction is \textit{valid} if its set of premises deductively implies the conclusion.}.
It is particularly central because making inferential commitments explicit both clarifies the underlying logic of an argument and enables subsequent argument evaluation by assessing the correctness of each premise~\citep{ct-3steps-4,walton2008enthymeme,baronett5e_reconstructing}.
We present an example of critical thinking via argument reconstruction in Figure~\ref{fig:aquinas}.

\begin{table*}[t]
\vspace{-5mm}
\centering
\scriptsize
\setlength{\tabcolsep}{3pt}
\renewcommand{\arraystretch}{1.15}

\begin{tabularx}{\textwidth}{l p{2cm} p{3.4cm} p{1.1cm} p{1.0cm} X}
\toprule
 & \textbf{Argument Length} & \textbf{Argument Types} & \textbf{Fallacy Inclusion} & \textbf{Domain} & \textbf{Argument Source} \\
\midrule
AAC, AAAC & Few sentences & Deduction & No & Logic & Symbolic engine-generation \\
EntailmentBank & Few sentences & Deduction (only 6 inference types) & No & QA & QA datasets \\
ARCHE & Single sentence & Deduction, Induction, Abduction & No & Science & Introduction of scientific papers \\
AAR & Few sentences & Deduction & No & Science & AI conference reviews \\
\midrule
\textbf{GAAR (Ours)} & Few sentences to Few paragraphs & (1) Deduction, Induction, Abduction, Analogy (general); (2) 60 informal logic schemes (specific) & Yes & General & News articles, Debate textbooks and websites, Argument evaluation datasets, LLM-generation \\
\bottomrule
\end{tabularx}

\caption{
Comparison of our argument reconstruction task with those in AAC~\citep{cript}, AAAC~\citep{deepa2}, EntailmentBank~\citep{entailmentbank}, ARCHE~\citep{arche}, and AAR~\citep{ryu2025reviewscore}.
}
\label{tab:dataset_comparison}
\vspace{-5mm}
\end{table*}

Following our three-stage operationalization of critical thinking, there is a line of work studies argument reconstruction with language models~\citep{cript, deepa2, entailmentbank, arche, ryu2025reviewscore}. However, these works only focus on reconstructing arguments that are relatively short and simple, mostly deductive, free of fallacies, and domain-specific (Table~\ref{tab:dataset_comparison}), and more importantly, these works do not investigate whether learning an argument reconstruction task improves performance on several critical thinking tasks.

To go beyond these preliminary approaches, we propose a holistic framework that systematically improves models' general critical thinking ability with argument reconstruction.
To this end, we (1) propose a generalized automatic argument reconstruction (GAAR) engine, (2) construct an argument reconstruction dataset (Arguinas), and (3) show that learning argument reconstruction benefits downstream critical thinking tasks.

First, we introduce GAAR, the first automatic engine that can \textit{faithfully} reconstruct arbitrary types of arguments\footnote{A reconstruction is \textit{faithful} if it preserves the intent of the argument without introducing or distorting information.} (Section~\ref{sec:gaar}).
To this end, we apply AAR~\citep{ryu2025reviewscore} to different types of arguments and observe two major limitations.
The first one is that AAR cannot faithfully reconstruct non-deductive or fallacious arguments. We address this limitation by teaching the model how to reconstruct arbitrary types of arguments, drawing from two argumentation theories~\citep{stanfordArgtype, walton2008argumentation}, and by embedding decision paths for reconstructing both formal and informal fallacies\footnote{An example of a formal fallacy is affirming the consequent: ``If it is snowing, then it is cold. It is cold. Therefore, it is snowing.'' An example of an informal fallacy is hasty generalization: ``His dog bit me, so all dogs are vicious.''}.
Another limitation is that AAR does not reliably judge the faithfulness of reconstructions because it relies on a coarse criterion. To address this issue, we propose three fine-grained criteria (i.e., \textit{accuracy}, \textit{completeness}, and \textit{parsimony}) that independently capture different aspects of faithfulness.
Our experimental results show that GAAR resolves these issues and outperforms all baseline methods, including AAR and LLM prompting, on reconstructing different types of arguments.

Second, we synthesize an \textbf{argu}ment reconstruct\textbf{i}o\textbf{n} dat\textbf{as}et (Arguinas) by running GAAR (Section~\ref{sec:dataset}).
We first collect arguments from seven sources to construct an argument text corpus.
We then select a base LLM to run the engine, choosing Claude Sonnet 4.5~\citep{sonnet4p5} as the most suitable model given the trade-off between reconstruction quality and API cost.
Finally, we define our argument reconstruction task as mapping an argument to a reconstruction.

Lastly, we empirically demonstrate that learning argument reconstruction benefits seven downstream critical thinking tasks (Section~\ref{sec:impact-on-critical}).
We finetune a pre-trained model on Arguinas and then further finetune the model on downstream tasks. The results show that our model outperforms the model directly finetuned on downstream tasks. Furthermore, training on Arguinas yields the largest performance gains compared to training on other argument reconstruction datasets (AAAC and EntailmentBank).
We also find that finetuning an instruction-tuned model on Arguinas also improves the model’s performance on downstream tasks, even without additional finetuning on the downstream tasks.
Furthermore, we analyze which components of Arguinas drive the downstream gains, providing a deeper understanding of how Arguinas is beneficial.

\begin{figure*}[t]
  \vspace{-8mm}
  \centering
  \includegraphics[width=\textwidth]{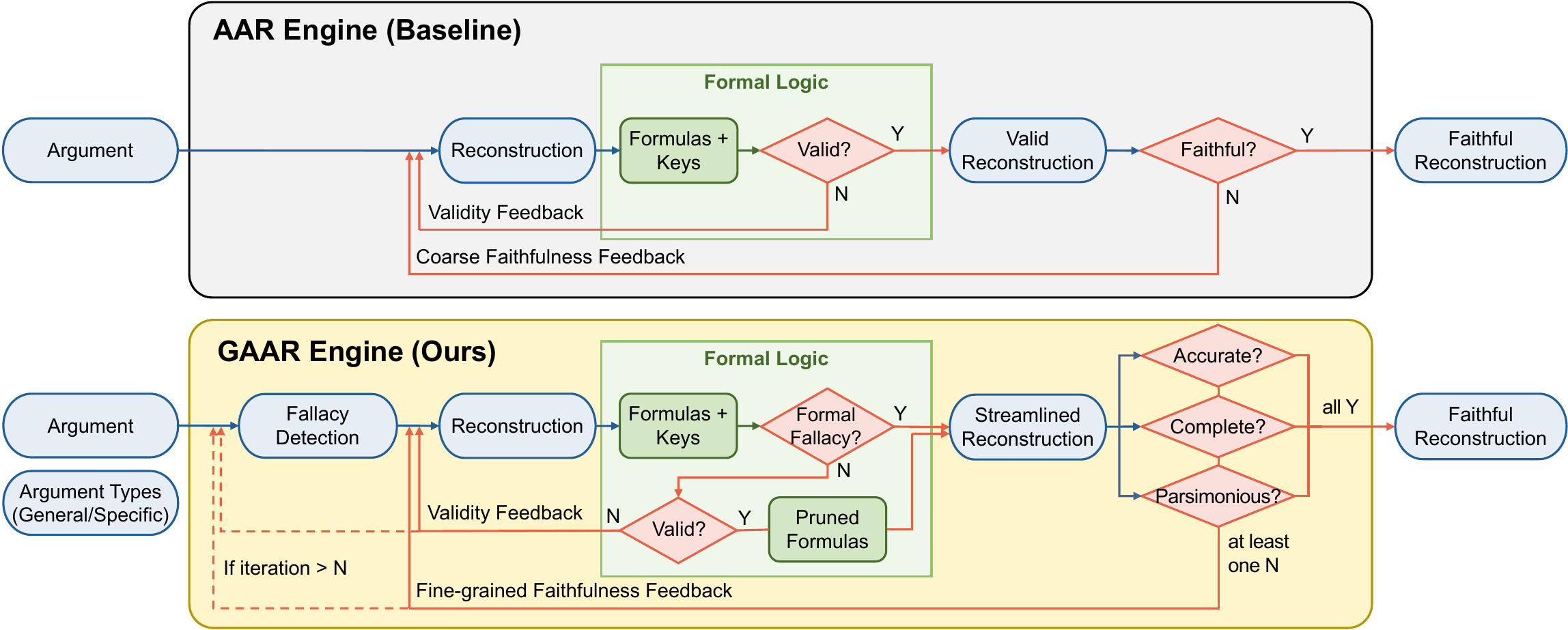}
  \caption{
  An overview of Generalized Automatic Argument Reconstruction (GAAR).
  }
  \label{fig:gaar}
  \vspace{-6mm}
\end{figure*}

\section{Related Works}
\label{sec:related-works}

Several works have studied how to reconstruct arguments using language models (Table~\ref{tab:dataset_comparison}). However, how these works define argument reconstruction tasks determines the required skill sets and their real-world applicability.
AAC~\citep{cript} and AAAC~\citep{deepa2} define the task as reconstructing an argument synthesized from symbolic inferences, so the primary required skill is identifying an argument structure, making these tasks less applicable.
EntailmentBank~\citep{entailmentbank} defines the task as reconstructing an argument with a conclusion and a set of sentences. Solving this task requires an additional skill---identifying explicit premises---but the task remains largely synthetic.
ARCHE~\citep{arche} defines the task as reconstructing a sentence from the introduction of a scientific paper, but it requires the same skill set as EntailmentBank.
Finally, AAR~\citep{ryu2025reviewscore} defines the task as reconstructing an AI conference review point. This task requires additional skills, including reformulating unclear and incomplete statements and identifying implicit premises~\citep{brun2016analysing}, making it the closest to the real-world application. However, the input arguments are assumed to be deductive and fallacy-free, which restricts the generalizability of the approach.
To resolve these issues, in this work, we define the task as reconstructing arbitrary arguments, which requires the full skills as AAR and is directly applicable to the real-world.


\begin{figure*}[t]
  \centering

  \begin{subfigure}[c]{0.5\textwidth}
    \centering
    \includegraphics[width=\linewidth]{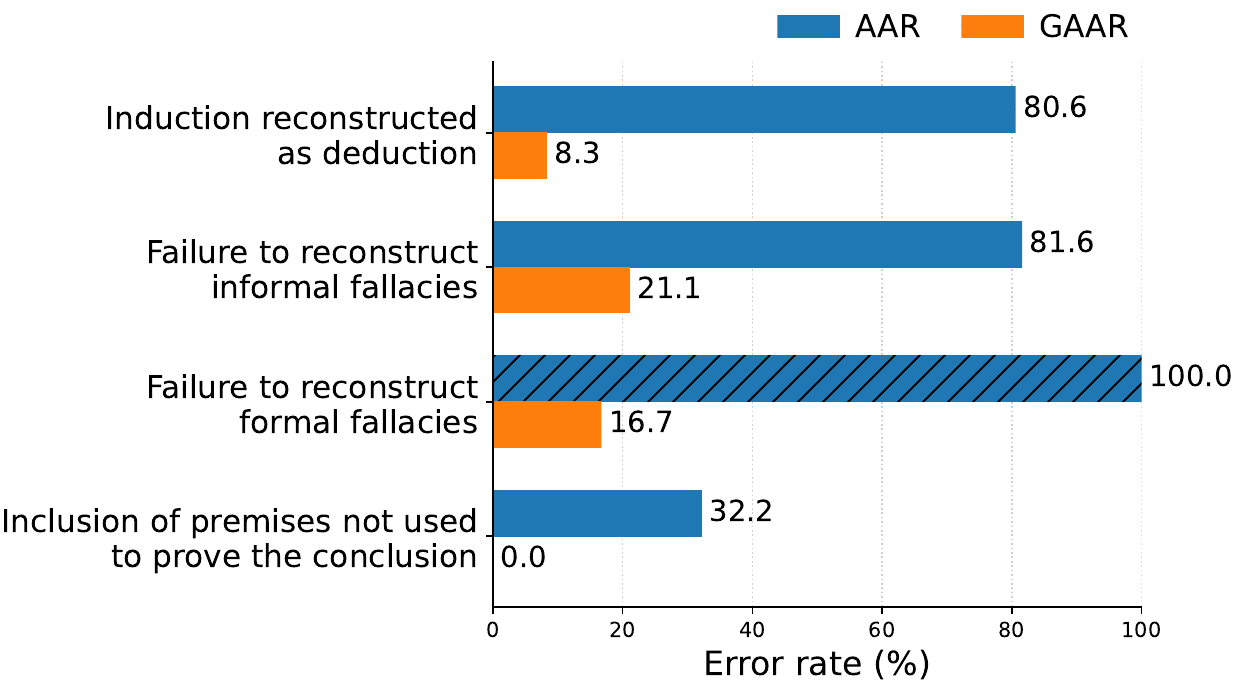}
    \vspace{-10mm}
    \caption{}
    \label{fig:aar-error-a}
  \end{subfigure}
  \hfill
  \begin{subfigure}[c]{0.49\textwidth}
    \centering
    \includegraphics[width=\linewidth]{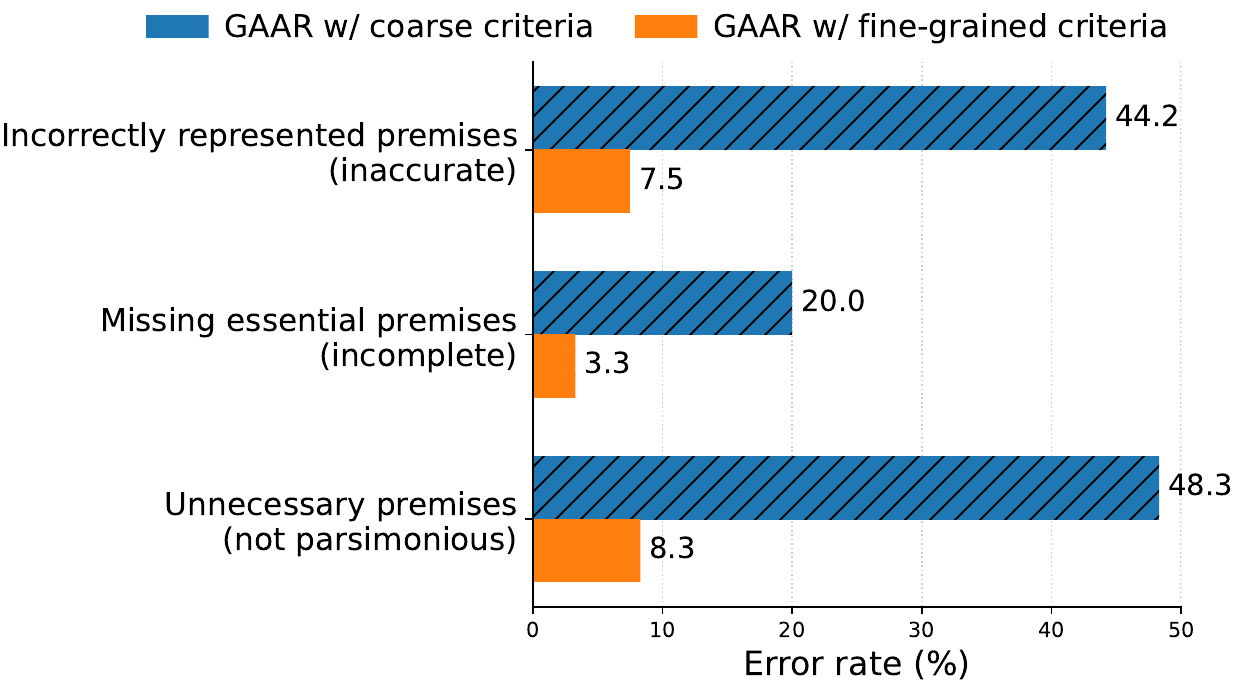}
    \vspace{-10mm}
    \caption{}
    \label{fig:aar-error-b}
  \end{subfigure}

  \caption{
    Quantitative analysis of the observed limitations of AAR and the improvements introduced by GAAR. To measure the error rates, we use GPT-5.2 to automatically evaluate each criterion on reconstructions whose source arguments are drawn from the Arguinas test set.
  }
  \label{fig:aar-error}
  \vspace{-5mm}
\end{figure*}

\vspace{-2mm}
\section{Generalized Automatic Argument Reconstruction}
\label{sec:gaar}
We propose Generalized Automatic Argument Reconstruction (GAAR), an automatic engine that can faithfully reconstruct arbitrary arguments into formal premise--conclusion structures.
To this end, we first briefly review the background of AAR (Section~\ref{subsec:aar-recap}) and then introduce GAAR by addressing the limitations of AAR (Section~\ref{subsec:gaar}).

\subsection{Background: AAR}
\label{subsec:aar-recap}
AAR~\citep{ryu2025reviewscore} is an automatic engine specifically designed to reconstruct argumentative AI conference review points (Figure~\ref{fig:gaar}).
Within the engine, an LLM first reconstructs an argument into a deductively valid premise--conclusion structure.
AAR then evaluates the generated reconstruction according to two criteria: the \textit{validity} of the inference and the \textit{faithfulness} of the premises and conclusion.
To verify \textit{validity}, an LLM translates the NL reconstruction (i.e., a set of premises and a conclusion) into corresponding first-order logic (FOL) formulas, after which a SAT solver determines whether the premises deductively imply the conclusion.
To verify \textit{faithfulness}, an LLM translates the FOL formulas back into NL premises and conclusion, and an LLM judge then determines whether the reconstruction faithfully represents the original argument. The reason the back-translated reconstruction is passed to the next stage instead of the initial reconstruction is to clarify the semantic content of the premises and how they logically relate to one another, a process known as \textit{logical streamlining}~\citep{deepa2}.
If either criterion is not satisfied, the corresponding feedback is sent back to the first stage to refine the reconstruction until both criteria are met.

\subsection{GAAR Overview}
\label{subsec:gaar}
We observe several limitations of AAR that prevent it from being generalized to arbitrary arguments. By addressing these issues, we propose GAAR, which can faithfully reconstruct arbitrary arguments (Figure~\ref{fig:gaar}).
In the following, we introduce four major improvements.
We also provide a stage-by-stage walkthrough in Appendix~\ref{app:sec:gaar} and \ref{app:sec:gaar_ex}.

\textbf{Arbitrary Argument Types.}
The most significant drawback of AAR is that it is optimized to reconstruct deductive arguments, which are prevalent in science domains but less common in other domains (e.g., political, societal, or legal). When AAR is applied to arguments from these domains, the original logical structures are often distorted. One prevailing pattern we observe is that AAR unfaithfully reconstructs 80.6\% of inductive arguments as deductive ones (Figure~\ref{fig:aar-error-a}).

To address this issue, we adopt multiple argument types from two argumentation theories and teach the model how to reconstruct each type of argument at the initial reconstruction stage.
The most widely-accepted, \textit{general} theory distinguishes four types of arguments: deduction, induction, abduction, and analogy~\citep{stanfordArgtype}. By definition, deductive arguments can be reconstructed in a deductively valid form (i.e., a set of premises deductively implies a conclusion). Moreover, prior works in informal logic~\citep{Feldman1993-FELRAA-5, stanfordInformalLogic} show that non-deductive arguments can also be reconstructed in a deductively valid form by adding the following connecting premise: ``If these premises are true, then the conclusion is true.''
Another, \textit{specific} theory~\citep{walton2008argumentation} further categorizes arguments into 60 types to fully capture everyday argumentation. Following the same justification, we confirm that Walton’s argument types can likewise be reconstructed in a deductively valid form.
The definitions of each argument type in both argumentation theories, along with their NL reconstructions, are provided in Appendix~\ref{app:sec:argtype}.

\textbf{Fallacy Inclusion.}
Another drawback of AAR is that it does not account for fallacious arguments.
Because AAR assumes all arguments can be reconstructed in a deductively valid form, arguments with formal fallacies cannot be faithfully reconstructed.
Although informal fallacies can, in principle, be reconstructed in a deductively valid form~\citep{barker1976fallacy}, we observe that AAR falls short of reconstructing 81.6\% of them (Figure~\ref{fig:aar-error-a}).

To address this issue, we incorporate decision logic for handling fallacies.
First, we introduce a fallacy detection stage prior to the initial reconstruction. This stage precedes argument reconstruction because the presence of fallacies strongly influences how an argument should be reconstructed, including whether the reconstruction should be deductively valid or not.
In the subsequent reconstruction stage, the model is provided with the detected fallacies along with their rationales, enabling it to reconstruct the argument accordingly. If the detection result is incorrect, the iterative loop would not terminate; therefore, we introduce two exceptional paths that allow the fallacy detection result to be revised after $N$ iterations.
Second, we add an additional exceptional path for arguments with formal fallacies that skips the validity judgment stage, since such arguments are invalid by definition.

\textbf{Fine-grained Faithfulness Evaluation Criteria.}
Now, the engine is designed to cover arbitrary arguments, including fallacious ones, but we still observe that the final reconstructed outputs are often unfaithful. Through manual inspection of the faithfulness judge’s decisions, we find that the evaluation criterion is overly coarse and frequently fails to detect undesirable patterns in the reconstructions.
We categorize such patterns into three types:
(1) incorrectly represented premises \textcolor{black}{(44.2\%)}, including overly generalized premises\footnote{e.g., an intended premise is ``This medicine reduces cold symptoms.'', but the reconstructed output contains ``All medicines reduce cold symptoms.''},
(2) missing premises that are essential for supporting the conclusion \textcolor{black}{(20.0\%)}, and
(3) unnecessary premises \textcolor{black}{(48.3\%)}, including rhetorical statements or paraphrased content \textcolor{black}{(Figure~\ref{fig:aar-error-b})}.

To address this issue, we propose three fine-grained criteria for evaluating the \textit{faithfulness} of reconstructions: \textit{accuracy}, \textit{completeness}, and \textit{parsimony}.
\textit{Accuracy} evaluates whether the premises and conclusion are precisely reconstructed without misinterpretation;
\textit{completeness} evaluates whether all necessary or core premises required to reconstruct an argument are included; and
\textit{parsimony} evaluates whether premises that are unnecessary for supporting the conclusion are excluded.
We replace the coarse criterion with these three fine-grained criteria at the faithfulness judging stage. If all three criteria are satisfied, the iterative loop terminates; otherwise, the corresponding fine-grained feedback is sent back to the reconstruction stage.

\textbf{Premise Pruning.}
An additional undesirable pattern is the inclusion of premises that are not used to prove the conclusion, \textcolor{black}{which occurs in 32.2\% of reconstructions (Figure~\ref{fig:aar-error-a})}.
To remove such premises, we employ a SAT solver to automatically perform premise pruning in the symbolic domain. Note that if there are multiple ways to prove the conclusion using different sets of premises, the SAT solver removes only those premises that are never used in any valid proof. With this premise-pruning mechanism integrated into the engine, a reconstruction that is judged to be valid is first pruned in the FOL domain and then translated back into the NL domain.

\section{Argument Reconstruction Dataset}
\label{sec:dataset}
Based on the GAAR engine, we synthesize an argument reconstruction dataset (Arguinas) (Section~\ref{subsec:data-synthesis}), verify its quality (Section~\ref{subsec:data-quality}), and analyze the dataset in several dimensions (Section~\ref{subsec:data-analysis}).

\begin{table}[t]
\centering
\scriptsize
\setlength{\tabcolsep}{2.5pt}
\renewcommand{\arraystretch}{1.15}

\begin{tabularx}{\columnwidth}{l c c c c c}
\toprule
\textbf{Sources} & \textbf{\# Data} & \textbf{\shortstack{\# Words\\in Arguments}} & \textbf{\shortstack{\# Premises}} & \textbf{\shortstack{\% Implicit\\Premises}} \\
\midrule
Procon.org                     & 282 & 178.01$\pm$100.68 & 6.46$\pm$2.95 & 45.39$\pm$19.39 \\
Pros-and-cons-1950             & 119 & 43.41$\pm$8.98    & 5.59$\pm$1.82 & 47.41$\pm$18.08 \\
Pros-and-cons-2010             & 373 & 83.53$\pm$25.81   & 5.79$\pm$2.20 & 51.99$\pm$18.13 \\
NYT-room-for-debate            & 297 & 398.21$\pm$92.02  & 8.13$\pm$3.39 & 41.68$\pm$15.54 \\
Anthropic-Persuasion           & 287 & 252.33$\pm$37.34  & 8.47$\pm$3.15 & 40.42$\pm$17.57 \\
Synthetic Arguments            & 1,520 & 332.60$\pm$190.55 & 9.14$\pm$4.28 & 37.88$\pm$15.29 \\
Synthetic Fallacies            & 297 & 296.45$\pm$160.88 & 8.40$\pm$4.28 & 35.70$\pm$19.11 \\
\midrule
Total                          & 3,175 & 269.48$\pm$177.21 & 8.15$\pm$3.94 & 40.94$\pm$17.46 \\
\bottomrule
\end{tabularx}
\caption{
Data statistics of arguments and their reconstructions in Arguinas.
\textcolor{black}{To measure the proportion of implicit premises, we prompt GPT-5.2 to classify each premise as either explicitly stated or implicitly assumed in the original argument.}
}
\label{tab:dataset_stats}
\vspace{-5mm}
\end{table}

\subsection{Data Synthesis}
\label{subsec:data-synthesis}
We run the GAAR engine to synthesize an argument reconstruction dataset.
To do so, we specify three components: (i) a target argument text corpus, (ii) a base LLM to run the engine, and (iii) a task definition that specifies the input--output format.

First, we collect \textcolor{black}{3,175} argument texts from seven different sources. These include two editions (1950 and 2010) of \emph{Pros and Cons: A Debater's Handbook}~\citep{pros-and-cons}, Britannica’s procon.org, the \emph{New York Times} column ``Room for Debate'', Anthropic-Persuasion~\citep{durmus2024measuring}, and LLM-generated arguments (both non-fallacious and fallacious).
\textcolor{black}{To synthesize non-fallacious arguments with LLMs, we collect debate topics from the args.me corpus~\citep{argsme} and prompt GPT-5 and GPT-5.1 to generate arguments supporting or opposing each topic. To synthesize fallacious arguments, we first generate additional arguments using the same procedure and then prompt GPT-5.2 to insert a specified type of formal or informal fallacy.}

Second, we select Claude Sonnet 4.5 as the base model. To make this choice, we run the GAAR engine using 13 frontier models and observe a clear trend: higher API cost generally corresponds to higher reconstruction quality. Given our limited resources, we balance cost and quality and conclude that Claude Sonnet 4.5 provides the best trade-off.

Finally, we formally define our reconstruction task as follows: the input consists of an argument and an instruction specifying argument types and their reconstructions, and the output is an argument reconstruction consisting of a list of premises and a conclusion. The exact prompt is shown in Figure~\ref{fig:prompt_recon_recon_llmprompting}.
Additional details of data synthesis process and data examples are provided in Appendix~\ref{app:sec:data-synthesis}.
\textcolor{black}{After synthesizing the Arguinas dataset, we split it into a training set of 2,934 samples and a test set of 241 samples while preserving the distribution of samples across argument data sources.}

\begin{table}[t]
\centering
\scriptsize
\setlength{\tabcolsep}{4pt}
\renewcommand{\arraystretch}{1.12}

\begin{tabularx}{0.85\columnwidth}{l c c}
\toprule
 & \textbf{Validity} & \textbf{Faithfulness}\\
\midrule
\textit{LLM Prompting}\\
gpt-oss-120b                  & 40.8 & 26.4 \\
Qwen3-235B-A22B-Thinking-2507 & 39.2 & 32.1 \\
Claude Sonnet 4.5 (no-think)  & 59.2 & 44.6 \\
GPT-5.2 (none)                & 51.7 & 46.2 \\
Claude Sonnet 4.5 (think)     & 60.8 & 48.0 \\
GPT-5.2 (xhigh)               & 80.8 & 48.8 \\
\midrule
\textit{LLM-based Engine}\\
AAR             & 100.0 & 21.4 \\
GAAR (general)  & 100.0 & N/A \\
GAAR (specific) & 100.0 & 46.5 \\
\bottomrule
\end{tabularx}
\caption{Argument reconstruction quality of GAAR and the baselines. We use Claude Sonnet 4.5 (no-think) as a base model to run LLM-based engines.
For faithfulness evaluation, we report the baselines’ winning rates against GAAR (general), excluding ties (\%).
\textcolor{black}{Here, "general" and "specific" in parentheses refer to two different theories of argument types.}}
\label{tab:gaar-quality}
\vspace{-5mm}
\end{table}

\subsection{Data Quality}
\label{subsec:data-quality}
We verify the quality of argument reconstruction data generated by GAAR.
Following AAR, we measure \textit{validity} and \textit{faithfulness} of reconstruction data.
To measure \textit{validity}, an LLM first translates the NL reconstruction into FOL formulas, after which a SAT solver determines whether the resulting proof is valid.
To measure \textit{faithfulness}, using an LLM judge with the fine-grained criteria, we report the baselines' winning rates against the proposed GAAR (general), excluding ties.
We use Claude Sonnet 4.5 to automate the evaluations.

\textcolor{black}{To verify the reliability of the automatic evaluations, we conduct human evaluations.
For \textit{validity}, because a SAT solver’s decision is mathematically correct, we evaluate the accuracy of the model’s NL-to-FOL translations. We randomly sample 209 NL-FOL pairs from the first and final turns of the GAAR engine, and a human expert manually determines whether each translation is correct.
For \textit{faithfulness}, we randomly sample 120 reconstruction pairs covering comparisons between GAAR (general) and three baselines: AAR, GAAR (specific), and Claude Sonnet 4.5 (think) prompting. A human expert then manually determines whether each LLM judge decision is correct.
The results show that the model achieves 99.0\% accuracy in NL-to-FOL translation and 89.2\% accuracy in pairwise faithfulness judgments.}

Evaluating on arguments in the Arguinas test set, we observe that the proposed GAAR outperforms the baseline AAR and LLM prompting methods both in validity and faithfulness (Table~\ref{tab:gaar-quality}). Specifically, GAAR and AAR shows perfect validity, whereas LLM prompting methods exhibit low validity. In terms of faithfulness, GAAR significantly outperforms AAR, which supports the effectiveness of our improvements in Section~\ref{subsec:gaar}.
We also conduct ablation studies, and the results show that GAAR outperforms any ablated method (Appendix~\ref{app:sec:gaar-ablation}).
\textcolor{black}{To make our automatic evaluation more reliable, we use two additional LLMs, GPT-5.1 and Gemini 3.1 Pro, as faithfulness judges. The results show that GAAR consistently outperforms the baselines across all judges (Appendix \ref{app:sec:arg-recon-eval-add}).}

\subsection{Data Analysis}
\label{subsec:data-analysis}
We analyze Arguinas with two dimensions, arguments and their reconstructions (Table~\ref{tab:dataset_stats}).
In terms of arguments, 42.8\% of the collected ones are written by humans. Other half of the arguments are written by LLMs (especially, GPT-5/5.1/5.2).
There are average 269.48 words in arguments but the number varies a lot depending on the source. For instance, arguments from Pros and cons (1950 edition) contain average 43.41 words, but arguments from New York Times column contain average 398.21 words, which is 9.2 times longer.
In terms of reconstructions, there are average 8.15 premises in a single argument. An interesting point is that arguments in New York Times column contain the largest number of words, but these contain smaller number of premises than synthetic arguments which contain smaller number of words. This suggests that humans include many rhetoric in their argument, whereas LLMs do not.
Also, among all explicit and implicit premises, the average portion of implicit premises is 40.94\%, which is almost comparable to that of explicit premises. Another interesting point is that LLMs use fewer implicit premises than humans, which means LLMs generate more direct arguments.

\begin{table*}[t]
\vspace{-5mm}
\centering
\scriptsize
\setlength{\tabcolsep}{4pt}
\renewcommand{\arraystretch}{1.15}

\begin{tabularx}{\textwidth}{l c c c c c c c}
\toprule
 & \multicolumn{3}{c}{Argument Quality Evaluation} & \multicolumn{2}{c}{Argument Reasoning} & Legal Reasoning & Logical Reasoning \\
\cmidrule(lr){2-4}\cmidrule(lr){5-6}\cmidrule(lr){7-7}\cmidrule(lr){8-8}
 & \text{\shortstack{WebisArgQuality20}} & \text{\shortstack{UKPConvArg2}} & \text{\shortstack{WebisCMV20}} & \text{\shortstack{ArgsNovel}} & \text{\shortstack{ArgRC}} & \text{\shortstack{LegalArg}} & \text{\shortstack{ReClor}}\\
\midrule
\multicolumn{8}{c}{\textit{Pre-adaptive Finetuning}} \\
\addlinespace[2pt]

\multicolumn{8}{l}{\textbf{Qwen3-4B-Base}} \\
Target-SFT                 & 39.37 & 92.35 & 53.80 & 51.99 & 86.54 & 47.24 & 74.13 \\
ArgumentOnly-Target-SFT    & \textbf{41.08} & 92.45 & 55.45 & 49.88 & 88.26 & 48.60 & 78.28 \\
UnstructuredAnalysis-Target-SFT & 34.77 & 91.21 & 49.95 & 49.24 & 90.20 & 51.70 & 74.25 \\
AAAC-Target-SFT            & 38.58 & 92.74 & 53.34 & 49.44 & 87.33 & 56.06 & 77.15 \\
EntailmentBank-Target-SFT  & 38.00 & 92.49 & 55.33 & 49.07 & 87.16 & 55.53 & 77.55 \\
\rowcolor{lightgray} Arguinas-Target-SFT (Ours)  & 41.01 & \textbf{92.79} & \textbf{56.03} & \textbf{52.52} & \textbf{90.77} & \textbf{58.52} & \textbf{78.98} \\

\addlinespace[3pt]
\multicolumn{8}{l}{\textbf{Qwen3-8B-Base}} \\
Target-SFT                 & 34.97 & 92.49 & 53.30 & 53.05 & 91.67 & 54.70 & 84.48 \\
ArgumentOnly-Target-SFT    & 35.27 & 92.56 & 50.57 & 46.35 & 88.57 & 57.89 & 83.70 \\
UnstructuredAnalysis-Target-SFT & 33.44 & 89.03 & 51.51 & 46.44 & 89.86 & 56.36 & 84.30 \\
AAAC-Target-SFT            & 37.67 & 92.12 & 51.46 & 49.85 & 90.68 & 61.63 & 84.70 \\
EntailmentBank-Target-SFT  & \textbf{38.96} & 91.40 & 50.22 & 45.98 & 90.63 & 61.24 & 84.05 \\
\rowcolor{lightgray} Arguinas-Target-SFT (Ours)  & 37.77 & \textbf{92.86} & \textbf{56.70} & \textbf{53.39} & \textbf{92.63} & \textbf{62.87} & \textbf{85.43} \\

\midrule
\multicolumn{8}{c}{\textit{Continued Finetuning}} \\
\addlinespace[2pt]

\multicolumn{8}{l}{\textbf{Qwen3-4B-Instruct}} \\
Off-the-shelf              & 34.68 & 77.66 & 52.50 & 49.30 & 71.68 & \textbf{65.42} & \textbf{81.30} \\
\rowcolor{lightgray} Arguinas-SFT (Ours)  & \textbf{36.43} & \textbf{80.22} & \textbf{53.42} & \textbf{53.59} & \textbf{73.76} & 61.75 & 81.05 \\

\addlinespace[3pt]
\multicolumn{8}{l}{\textbf{Qwen2.5-7B-Instruct}} \\
Off-the-shelf              & 22.35 & 77.45 & 50.63 & \textbf{49.72} & 70.02 & 60.19 & 69.40 \\
\rowcolor{lightgray} Arguinas-SFT (Ours)  & \textbf{33.82} & \textbf{77.59} & \textbf{52.58} & 47.52 & \textbf{74.32} & \textbf{61.67} & \textbf{73.05} \\

\bottomrule
\end{tabularx}
\caption{Downstream critical thinking task performance of our finetuning strategies (\textit{pre-adaptive} and \textit{continued} finetuning) and the baseline methods. Macro F1 scores (\%) are reported for WebisArgQuality20, ArgsNovel, and LegalArg; accuracy (\%) is reported for the remaining tasks.
}
\label{tab:downstream-ft}
\vspace{-5mm}
\end{table*}

\section{Impact of Learning Argument Reconstruction on Critical Thinking Tasks}
\label{sec:impact-on-critical}
Finally, we demonstrate that learning argument reconstruction improves models’ general critical thinking ability.
We consider two scenarios: \textit{pre-adaptive} finetuning and \textit{continued} finetuning on the proposed Arguinas dataset.
We define finetuning on Arguinas as \textit{pre-adaptive} when it is used as an intermediate stage before downstream task finetuning (Pre-trained model $\rightarrow$ Arguinas $\rightarrow$ Downstream), and as \textit{continued} when it further finetunes an already instruction-tuned model without an additional downstream stage (Instruct model $\rightarrow$ Arguinas).
In the following, we introduce downstream critical thinking tasks (Section~\ref{subsec:downstream}) and present experiments and results of the two finetuning scenarios\footnote{For model finetuning, we exclude fallacious arguments from the Arguinas dataset since these have a detrimental effect on downstream (See Section~\ref{subsec:analysis}).} (Section~\ref{subsec:main-results}). Furthermore, we analyze which components of Arguinas drive the downstream gain (Section~\ref{subsec:analysis}).

\subsection{Critical Thinking Tasks}
\label{subsec:downstream}
\textcolor{black}{Under our operationalization, critical thinking consists of argument identification, reconstruction, and evaluation. The seven selected downstream tasks instantiate these core skills in different forms.
WebisArgQuality20~\citep{webisargqulaity20}, UKPConvArg2~\citep{ukpconvarg2}, and WebisCMV20~\citep{webiscmv20} directly require evaluating argument quality;
ArgsNovel~\citep{argvalidity} requires determining whether a conclusion is novel relative to its premises, which implicitly requires reconstructing the argument;
ArgRC~\citep{argrc} directly tests the reconstruction of implicit warrants;
LegalArg~\citep{legalarg} requires determining whether a legal argument is correct, which involves both reconstructing and evaluating arguments; and ReClor~\citep{reclor} requires examining, analyzing, and evaluating arguments.}
Detailed dataset statistics are described in Appendix~\ref{app:sec:downstream-data-stat}.

\subsection{Main Results}
\label{subsec:main-results}
\textcolor{black}{
\textbf{Experimental Setup.}
We perform completion-only supervised finetuning, except when finetuning on argument texts only. At each finetuning stage, we train for 3 epochs using fixed learning rate and batch size pairs according to the size of the training set (Table~\ref{tab:batch_lr}).
For a fair comparison with the baselines, we equalize the total number of training steps.
To evaluate finetuned model's downstream performance, we generate eight responses with temperature 0.7, top-p 0.8, and top-k 20 and report the averaged performance. We report Macro F1 for WebisArgQuality20, ArgsNovel, and LegalArg, and Accuracy for all other tasks.}

\textbf{Pre-adaptive Finetuning.}
In this scenario, we finetune a pre-trained model on Arguinas training set and then further finetune it on downstream tasks. We evaluate the resulting model on the downstream tasks and compare its performance with that of a model that is directly finetuned on the downstream tasks. We also include baselines such as replacing Arguinas with other argument reconstruction datasets or finetuning only on the argument texts from Arguinas. To verify the effectiveness of our structured argument analysis, we add a baseline that replaces Arguinas with an unstructured analysis dataset constructed by prompting an LLM\footnote{We prompt Qwen3-235B-A22B-Thinking-2507 to analyze a given argument within 250 words, which matches the average length of argument reconstruction data in Arguinas.}.
We use Qwen3-4B-Base and Qwen3-8B-Base~\citep{qwen3} as the initial pre-trained models.
Experimental results show that finetuning on Arguinas outperforms baselines on downstream tasks, except for WebisArgQuality20, where our method achieves the second-best performance (Table~\ref{tab:downstream-ft}).
We also observe that finetuning on other argument reconstruction tasks does not consistently benefit downstream performance, as these tasks are less representative of real-world argument analysis.
Furthermore, we use Llama-3.1-8B-Base~\citep{llama3} as the initial model and observe a similar trend (Appendix~\ref{app:sec:llama}).

\begin{figure*}[t]
  \vspace{-5mm}
  \centering

  \begin{subfigure}[c]{0.82\textwidth}
    \centering
    \includegraphics[width=\linewidth]{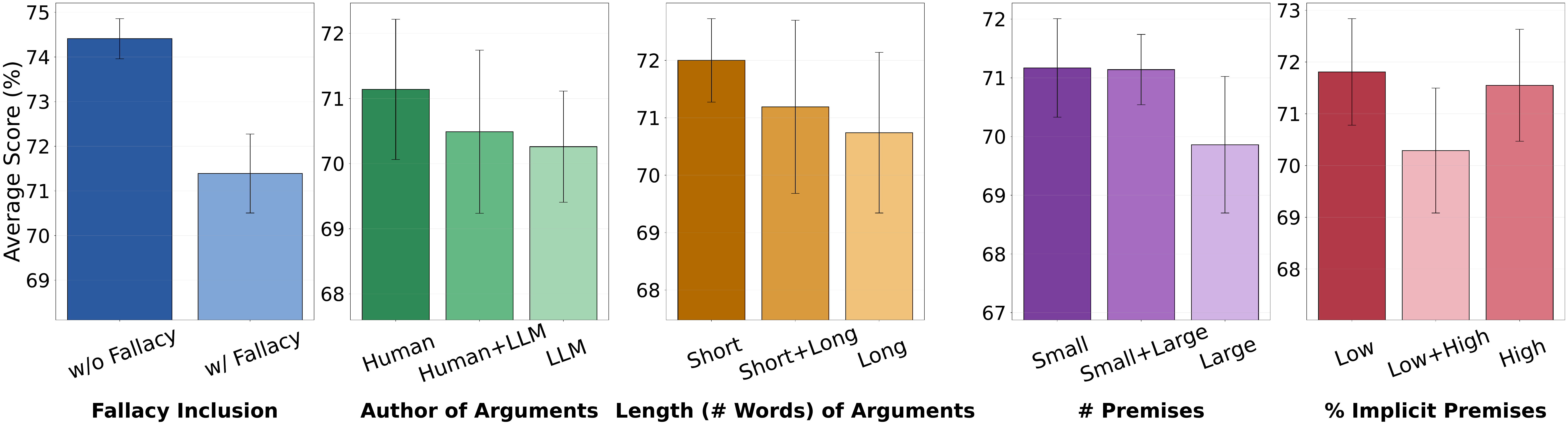}
    \vspace{-8mm}
    \caption{}
    \label{fig:arguinas-data-composition}
  \end{subfigure}
  \hfill
  \begin{subfigure}[c]{0.167\textwidth}
    \centering
    \includegraphics[width=\linewidth]{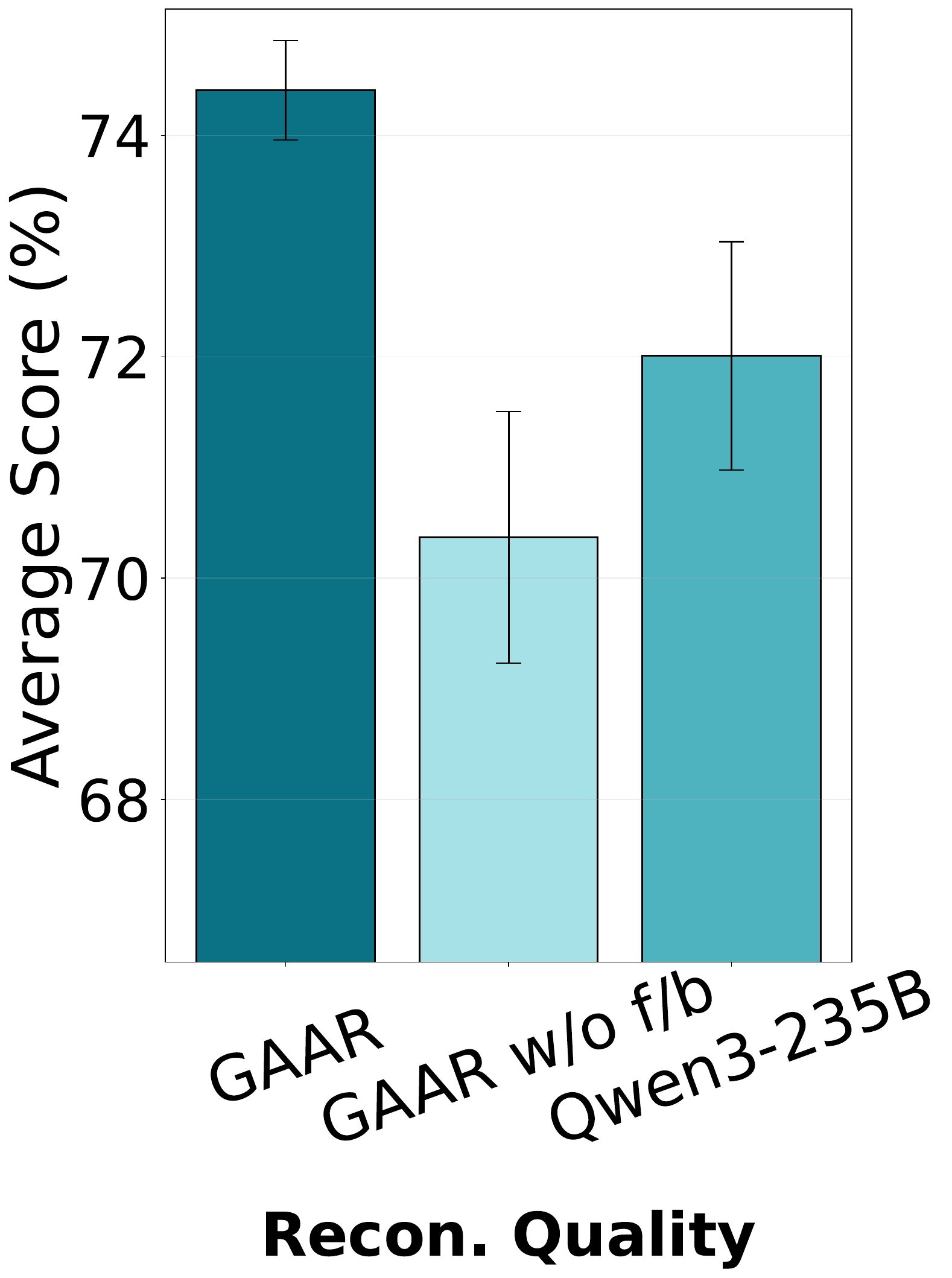}
    \vspace{-8mm}
    \caption{}
    \label{fig:recon-quality-downstream-impact}
  \end{subfigure}
  \vspace{-2mm}
  \caption{
    Average downstream task performance under the pre-adaptive finetuning
    scenario: (a) performance with different Arguinas data compositions and
    (b) performance with different argument reconstruction qualities.
  }
  \label{fig:arguinas-analysis}
  \vspace{-5mm}
\end{figure*}

\textbf{Continued Finetuning.}
In this scenario, we finetune an instruction-tuned model on Arguinas training set without further finetuning on downstream tasks. We evaluate the resulting model on downstream tasks and compare its performance with that of the original instruction-tuned model. We use Qwen3-4B-Instruct-2507~\citep{qwen3} and Qwen2.5-7B-Instruct~\citep{qwen2p5} as the instruction-tuned models.
Results show that continued finetuning on Arguinas generally improves the performance of instruction-tuned models on downstream tasks (Table~\ref{tab:downstream-ft}). The largest improvement is observed on WebisArgQuality20 using Qwen2.5-7B-Instruct, where F1 score is increased by 51.3\%.

In addition, we evaluate the finetuned models on the argument reconstruction task (Arguinas) and report the results in Appendix~\ref{app:sec:eval-on-arguinas}.
Furthermore, to support the statistical significance of downstream gains through finetuning on Arguinas, we run paired bootstrap tests in Appendix~\ref{app:sec:stat-signif} and show that most improvements are statistically significant.

\subsection{Analysis}
\label{subsec:analysis}

\textbf{Impact of Arguinas Data Composition.}
To understand which components of Arguinas contribute to downstream performance gains, we construct subsets of Arguinas by controlling five dimensions and measure the downstream impact of each subset (Figure~\ref{fig:arguinas-data-composition}).
Following the dimensions in Table~\ref{tab:dataset_stats}, we consider fallacy inclusion, author of arguments, argument length (number of words), number of premises, and the ratio of implicit premises.
We use Qwen3-8B-Base as the initial model and evaluate on four downstream tasks (WebisCMV20, ArgRC, LegalArg, and ReClor). For fair comparison, we equalize the subset size across each dimension\footnote{Detailed experimental setup is provided in Appendix~\ref{app:subsec:data-stat-arguinas-subset}.}.

First, learning to reconstruct fallacious arguments has a detrimental effect on downstream performance. This is because arguments in the downstream tasks are mostly designed to be free of fallacies. Based on this observation, we exclude fallacious arguments from Arguinas during model finetuning.
Second, human-written arguments yield the most positive impact on downstream performance, whereas incorporating LLM-generated arguments is detrimental.
However, since the performance gap between human-only and mixed subsets is marginal, we include LLM-generated arguments in finetuning to scale the size of the Arguinas dataset.
Next, long arguments (290+ words) and arguments with a large number of premises (9+ premises) degrade downstream performance. This is likely because most downstream arguments are short (only 10.8\% are long) and therefore contain fewer premises.
Finally, the effect of the ratio of implicit premises does not show a consistent trend and depends on the nature of the downstream task. For ArgRC, where the task is to identify a correct implicit premise, a high ratio of implicit premises is most beneficial. In contrast, for ReClor, where premises are mostly explicitly stated, a low ratio of implicit premises is most beneficial.

\textbf{Impact of Argument Reconstruction Quality.}
To verify that improvements in argument reconstruction quality contribute to downstream gains, we construct two lower-quality argument reconstruction datasets and compare their downstream effects with our method (Figure~\ref{fig:recon-quality-downstream-impact}).
Specifically, we collect one dataset from the GAAR engine before applying any faithfulness feedback, and another by prompting Qwen3-235B-A22B-Thinking-2507.
We use Qwen3-8B-Base as the initial model and evaluate on four downstream tasks (WebisCMV20, ArgRC, LegalArg, and ReClor). For fair comparison, we control for dataset size across all settings.
The results show that our GAAR-based method yields the largest downstream gains compared to both lower-quality datasets.

\textbf{Data Efficiency.}
\textcolor{black}{In addition, we analyze whether pre-adaptation on Arguinas reduces the amount of task-specific supervision required to achieve strong downstream performance (Figure~\ref{fig:data-efficiency}). We use three downstream tasks (UKPConvArg2, WebisCMV20, and ReClor) and observe a clear trend that the model finetuned on Arguinas requires significantly less downstream training data to achieve a similar downstream performance compared to the model not finetuned on Arguinas.}

\section{Conclusion}
We propose a holistic framework that enhances LLMs’ critical thinking ability through argument reconstruction.
We first develop an automatic engine, GAAR, that faithfully reconstructs arbitrary arguments.
Building on this engine, we synthesize a high-quality argument reconstruction dataset, Arguinas.
Finally, through systematic experiments, we empirically demonstrate that finetuning on Arguinas improves models’ performance on downstream critical thinking tasks.

\section*{Limitations}
\textcolor{black}{We acknowledge two limitations of our work.}

\textcolor{black}{First, when synthesizing the Arguinas dataset, we use only a single base LLM to run the GAAR engine, which may introduce model-specific biases. One way to mitigate this issue would be to run GAAR with multiple base models and select the best reconstruction for inclusion in the dataset. However, given our limited budget, we leave this extension to future work.}

\textcolor{black}{Second, because an argument may admit multiple faithful reconstructions, supervised finetuning on a single reference reconstruction may not be the optimal way to teach models how to reconstruct arguments. An alternative approach would be to develop a reconstruction evaluator and use it to train a policy model through reinforcement learning, with reconstruction quality serving as the reward signal. A key challenge, however, is building a sufficiently reliable reconstruction evaluator. We therefore leave this direction to future work as well.}

\section*{Acknowledgments}
This work was supported by Institute of Information \& communications Technology Planning \& Evaluation (IITP) grant funded by the Korea government (MSIT) (No. 2022-0-00984, Development of Artificial Intelligence Technology for Personalized Plug-and-Play Explanation and Verification of Explanation; No. RS-2024-00457882, AI Research Hub Project; RS-2022-00143911, AI Excellence Global Innovative Leader Education Program) and National Research Foundation of Korea (NRF) grant (No.RS-2023-00209060, A Study on Optimization and Network Interpretation Method for Large-Scale Machine Learning) funded by the Korea government (MSIT).

\bibliography{custom}

\clearpage

\appendix

\section{Types of Arguments and Their Reconstructions}
\label{app:sec:argtype}

In Section~\ref{subsec:gaar}, to reconstruct arbitrary arguments, we adopt two argumentation theories that propose different types of arguments.
One is the \textit{general} theory where it distinguishes four types of arguments: deduction, induction, abduction, and analogy~\citep{stanfordArgtype}. In Figure~\ref{fig:argtype_classic}, we present the definition of four argument types and how to reconstruct these.
Another is the \textit{specific} theory developed by Walton that we could distinguish 60 types (or schemes) of everyday arguments~\citep{walton2008argumentation}. In Figure~\ref{fig:argtype_walton_1}--\ref{fig:argtype_walton_3}, we present a full list of Walton's 60 arguments. We also include reconstructions of the first eight types in Figure~\ref{fig:argtype_walton_ex}. To reconstruct it in a deductively valid form, we simply add a conditional premise for each argument type. Full reconstructions of 60 argument types are available in Walton's original work~\citep{walton2008argumentation}.

\begin{figure}[h]
\footnotesize
\begin{tcblisting}{text only,
    halign=left, 
    title=\textbf{General Argument Types},
    colbacktitle=gray!30!white, 
    coltitle=black,
}
1. Deductive Reasoning: A form of argument where the truth of the premises necessitates the truth of the conclusion; the conclusion cannot but be true if the premises are true.\\
~\\

2. Inductive Reasoning: A form of ampliative argument where observations about past instances and regularities lead to generalizations about future instances and universal principles.\\
\vspace{5pt}
\textit{**Formalization**}\\
Premise\_1: Evidence\_1\\
Premise\_2: Evidence\_2\\
...\\
Premise\_N: Evidence\_N\\
Premise\_N+1: If Evidence\_1, Evidence\_2, ..., and Evidence\_N, then Generalization\\
$\therefore$~Conclusion: Generalization\\
~\\

3. Analogical Reasoning: A form of argument based on similarity relations, where if the source domain and target domain are similar in certain known respects, and the source domain possesses a further feature, then the target domain is inferred to also have that feature or a similar counterpart.\\
\vspace{5pt}
\textit{**Formalization**}\\
Premise\_1: Source S has a property P1\\
Premise\_2: Target T has a property P1* equal or similar to P1\\
Premise\_3: Source S has a property P2\\
Premise\_4: Target T has a property P2* equal or similar to P2\\
...\\
Premise\_(2N-1): Source S has a property Pn\\
Premise\_2N: Target T has a property Pn* equal or similar to Pn\\
Premise\_(2N+1): Source S has a property Q\\
Premise\_(2N+2): If source S has properties P1, P2, ..., Pn, target T has properties P1*, P2*, ..., Pn*, and source S has a property Q, then target T has a property Q* equal or similar to Q\\
$\therefore$~Conclusion: Target T has a property Q* equal or similar to Q\\
~\\

4. Abductive Reasoning: A form of ampliative argument that involves inference to the best explanation, where a conclusion is drawn as to what could plausibly explain the occurrence of observed facts.\\
\vspace{5pt}
\textit{**Formalization**}\\
Premise\_1: Observation\_1\\
Premise\_2: Observation\_2\\
...\\
Premise\_N: Observation\_N\\
Premise\_N+1: Explanation H explains Observation\_1, Observation\_2, ..., and Observation\_N\\
Premise\_N+2: If Observation\_1, Observation\_2, ..., and Observation\_N, and (Explanation H explains Observation\_1, Observation\_2, ..., and Observation\_N), then Explanation H is True\\
$\therefore$~Conclusion: Explanation H is True\\
\end{tcblisting}
\vspace{-2mm}
\caption{General Argument Types and Their Reconstructions.}
\label{fig:argtype_classic}
%
\end{figure}

\begin{figure}[t]
\footnotesize
\begin{tcblisting}{text only,
    halign=left, 
    title=\textbf{Specific Argument Types},
    colbacktitle=gray!30!white, 
    coltitle=black,
}
1. Argument from Position to Know\\
2. Argument from Expert Opinion\\
3. Argument from Witness Testimony\\
4. Argument from Popular Opinion\\
Subtypes of 4. Argument from Popular Opinion:\\
4.1. Pop Scheme\\
4.2. Position-to-Know Ad Populum Argument\\
4.3. Expert Opinion Ad Populum Argument\\
4.4. Deliberation Ad Populum Argument\\
4.5. Moral Justification Ad Populum Argument\\
4.6. Moral Justification (Excuse Subtype) Ad Populum Argument\\
4.7. Snob Appeal Ad Populum Argument\\
4.8. Appeal to Vanity Ad Populum Argument\\
4.9. Rhetoric of Belonging Ad Populum Argument\\
4.10. Common Folks Ad Populum Argument\\
4.11. Common Folks (Group Subtype) Ad Populum Argument\\
5. Argument from Popular Practice\\
6. Argument from Example\\
6.1. Argument from Example\\
6.2. Argument from Illustration\\
6.3. Argument from Model\\
6.4. Argument from Anti-Model\\
7. Argument from Analogy\\
8. Practical Reasoning from Analogy\\
8.1. Positive Schema\\
8.2. Negative Schema\\
9. Argument from Composition\\
9.1. Generic Composition\\
9.2. Inclusion of the Part in the Whole\\
10. Argument from Division\\
10.1. Generic Division\\
10.2. Division of the Whole into its Parts\\
11. Argument from Oppositions\\
11.1. Descriptive Schemes\\
11.2. Normative Schemes\\
12. Rhetorical Argument from Oppositions\\
12.1. Normative Schemes\\
12.2. Descriptive Schemes\\
13. Argument from Alternatives\\
13.1. Cognitive Schemes\\
13.2. Normative Schemes\\
14. Argument from Verbal Classification\\
15. Argument from Definition to Verbal Classification\\
16. Argument from Vagueness of a Verbal Classification\\
17. Argument from Arbitrariness of a Verbal Classification\\
18. Argumentation from Interaction of Act and Person\\
18.1. Variant 1\\
18.2. Variant 2\\
19. Argumentation from Values\\
19.1. Variant 1: Positive Value\\
19.2. Variant 2: Negative Value\\
20. Argumentation from Sacrifice\\
\end{tcblisting}
\vspace{-2mm}
\caption{Specific Argument Types.}
\label{fig:argtype_walton_1}
%
\end{figure}

\begin{figure}[t]
\footnotesize
\begin{tcblisting}{text only,
    halign=left, 
    title=\textbf{Specific Argument Types},
    colbacktitle=gray!30!white, 
    coltitle=black,
}
21. Argumentation from the Group and Its Members\\
21.1. Variant 1\\
21.2. Variant 2\\
22. Practical Reasoning\\
22.1. Practical Inference\\
22.2. Necessary Condition Schema\\
22.3. Sufficient Condition Schema\\
22.4. Value-Based Practical Reasoning\\
22.5. Argument from Goal\\
22.6. Argumentation from Ends and Means\\
23. Two-Person Practical Reasoning\\
24. Argument from Waste\\
25. Argument from Sunk Costs\\
26. Argument from Ignorance\\
26.1. Negative Reasoning from Normal Expectations\\
26.2. Negative Practical Reasoning\\
27. Epistemic Argument from Ignorance\\
28. Argument from Cause to Effect\\
29. Argument from Correlation to Cause\\
30. Argument from Sign\\
31. Abductive Argumentation Scheme\\
31.1. Backward Argumentation Scheme\\
31.2. Forward Argumentation Scheme\\
31.3. Abductive Scheme for Argument from Action to Character\\
31.4. Scheme for Argument from Character to Action (Predictive)\\
31.5. Retroductive Scheme for Identifying an Agent from a Past Action\\
32. Argument from Evidence to a Hypothesis\\
32.1. Argument from Verification\\
32.2. Argument from Falsification\\
33. Argument from Consequences\\
33.1. Argument from Positive Consequences\\
33.2. Argument from Negative Consequences\\
33.3. Reasoning from Negative Consequences\\
33.4. Argument from Negative Consequences (Prudential Inference)\\
34. Pragmatic Argument from Alternatives\\
35. Argument from Threat\\
35.1. Argument from Disjunctive Ad Baculum Threat\\
36. Argument from Fear Appeal\\
37. Argument from Danger Appeal\\
38. Argument from Need for Help\\
39. Argument from Distress\\
40. Argument from Commitment\\
\end{tcblisting}
\vspace{-2mm}
\caption{Specific Argument Types.}
\label{fig:argtype_walton_2}
%
\end{figure}

\begin{figure}[t]
\footnotesize
\begin{tcblisting}{text only,
    halign=left, 
    title=\textbf{Specific Argument Types},
    colbacktitle=gray!30!white, 
    coltitle=black,
}
41. Ethotic Argument\\
42. Generic Ad Hominem\\
43. Pragmatic Inconsistency\\
44. Argument from Inconsistent Commitment\\
45. Circumstantial Ad Hominem\\
46. Argument from Bias\\
47. Bias Ad Hominem\\
48. Argument from Gradualism\\
49. Slippery Slope Argument\\
50. Precedent Slippery Slope Argument\\
51. Sorites Slippery Slope Argument\\
52. Verbal Slippery Slope Argument\\
53. Full Slippery Slope Argument\\
54. Argument for Constitutive-Rule Claims\\
54.1. Physical World Premise Version 1\\
54.2. Physical World Premise Version 2\\
54.3. Mental World Premise\\
55. Argument from Rules\\
55.1. From Established Rule\\
55.2. From Rules\\
55.3. Regulative-Rule Premise Obligation Claim\\
56. Argument for an Exceptional Case\\
57. Argument from Precedent\\
58. Argument from Plea for Excuse\\
59. Argument from Perception\\
59.1. Argument from Perception\\
59.2. Argument from Appearance\\
60. Argument from Memory\\
\end{tcblisting}
\vspace{-2mm}
\caption{Specific Argument Types.}
\label{fig:argtype_walton_3}
%
\end{figure}

\begin{figure}[h]
\footnotesize
\begin{tcblisting}{text only,
    halign=left, 
    title=\textbf{Specific Argument Types},
    colbacktitle=gray!30!white, 
    coltitle=black,
}
1. Argument from Position to Know\\
Major Premise: Source a is in position to know about things in a certain subject domain S containing proposition A.\\
Minor Premise: a asserts that A is true (false).\\
Conclusion: A is true (false).\\
~\\

2. Argument from Expert Opinion\\
Major Premise: Source E is an expert in subject domain S containing proposition A.\\
Minor Premise: E asserts that proposition A is true (false).\\
Conclusion: A is true (false).\\
~\\

3. Argument from Witness Testimony\\
Position to Know Premise: Witness W is in a position to know whether A is true or not.\\
Truth Telling Premise: Witness W is telling the truth (as W knows it).\\
Statement Premise: Witness W states that A is true (false).\\
Conclusion: A may be plausibly taken to be true (false).\\
~\\

4. Argument from Popular Opinion\\
General Acceptance Premise: A is generally accepted as true.\\
Presumption Premise: If A is generally accepted as true, that gives a reason in favor of A.\\
Conclusion: There is a reason in favor of A.\\
~\\

5. Argument from Popular Practice\\
Major Premise: A is a popular practice among those who are familiar with what is acceptable or not in regard to A.\\
Minor Premise: If A is a popular practice among those familiar with what is acceptable or not with regard to A, that gives a reason to think that A is acceptable.\\
Conclusion: Therefore, A is acceptable in this case.\\
~\\

6. Argument from Example\\
Premise: In this particular case, the individual a has property F and also property G.\\
Conclusion: Therefore, generally, if x has property F, then it also has property G.\\
~\\

7. Argument from Analogy\\
Similarity Premise: Generally, case C1 is similar to case C2.\\
Base Premise: A is true (false) in case C1.\\
Conclusion: A is true (false) in case C2.\\
~\\

8. Practical Reasoning from Analogy (Positive Schema)\\
Base Premise: The right thing to do in S1 was to carry out action x.\\
Similarity Premise: S2 is similar to S1.\\
Conclusion: Therefore, the right thing to do in S2 is carry out x.\\
\end{tcblisting}
\vspace{-2mm}
\caption{Specific Argument Types and Their Reconstructions.}
\label{fig:argtype_walton_ex}
%
\end{figure}

\clearpage

\section{Detailed Description of GAAR Stages}
\label{app:sec:gaar}

In Section~\ref{subsec:gaar}, we describe GAAR by how it resolves existing issues of AAR.
To those who are not familiar with AAR, we prepare a complementary section that does not require prior knowledge of AAR where we describe GAAR by how each stage is operating.

Our engine consists of six stages, and we discuss how each stage operates in detail.
\paragraph{Stage 1. Fallacy Detection.}
Before an LLM reconstructs an argument, the model first detects if there are any (formal or informal) fallacy in the argument.
The reason why this stage precedes the reconstruction stage is that inference patterns in the reconstruction completely differ depending on whether the argument contains a fallacy. Specifically, arguments without formal fallacies should be reconstructed in a deductively valid form, but arguments with formal fallacies should not be. We also include detecting informal fallacies to facilitate subsequent reconstruction process.

\paragraph{Stage 2. Argument Reconstruction.}
An LLM reconstructs an argument into a set of NL premises and conclusion. Along with the argument, the model is also given argument types and their reconstructions (of either \textit{general} or \textit{specific} argumentation theory) and the fallacy detection result, which enables reconstructing arbitrary arguments.

\paragraph{Stage 3. Formalization.}
An LLM translates NL premises and conclusion into first-order logic (FOL) formulas. For later back-translation, the model also generates keys that map FOL predicates and/or variables to NL phrases.

\paragraph{Stage 4. Validity Judgment \& Pruning.}
Based on the formalization, a SAT solver automatically determines whether the reconstruction is valid (i.e., the premises deductively imply the conclusion).
If the reconstruction is invalid, then this signal is sent to Stage 2 to refine the reconstruction.
Otherwise (i.e., the reconstruction is valid), a SAT solver removes premises that are unnecessary to prove the conclusion. If there are multiple ways to prove the conclusion using different subsets of premises, our engine only removes premises that never appear in any proof. We call this step as a pruning.
Note that this stage is only applied for arguments without formal fallacies.

\paragraph{Stage 5. Streamlining.}
Before the engine enters the final faithfulness judgment stage, an LLM back-translates FOL formulas into NL premises and conclusion based on the previously generated keys.
The reason why we use this back-translated reconstruction instead of the initial reconstruction in Stage 2 is to clarify the semantic meaning of premises in Stage 2 and how those logically relates one another. This process is called a \textit{logical streamlining}~\citep{deepa2}, and we call the back-translated reconstruction as a streamlined reconstruction.

\paragraph{Stage 6. Faithfulness Judgment \& Feedback.}
An LLM judge decides if the (streamlined) premises and conclusion are \textit{accurate}, \textit{complete}, and \textit{parsimonious}.
To do that, the judge model is given a detailed instruction including definition of these three criteria.
If all criteria are satisfied, then the loop ends. Otherwise, a corresponding feedback is sent to Stage 2 to refine the reconstruction.

An example of outputs at each stage is demonstrated in Appendix~\ref{app:sec:gaar_ex}.
All prompts that we use in each stage are included in Appendix~\ref{app:sec:prompts}.

\section{Stage-by-Stage Walkthrough of GAAR}
\label{app:sec:gaar_ex}
To facilitate a better understanding of the internal mechanics of the GAAR engine, we demonstrate example outputs at each stage.

\textbf{Argument}\mbox{}\\
\noindent
\begin{minipage}[t]{\linewidth}
\scriptsize
We allow contraception. Abortion is, in effect, no different – the prevention of the development of a potential human being. In the case of the `morning after pill', the analogy is even closer. If we allow these measures, then we should also allow abortion.
\end{minipage}
\vspace{1mm}

\textbf{Stage 1-1. Fallacy Detection}\mbox{}\\
\noindent
\begin{minipage}[t]{\linewidth}
\scriptsize
Informal fallacy of false equivalence is detected. (Not a formal fallacy)
\end{minipage}
\vspace{1mm}

\textbf{Stage 1-2. Reconstruction}\mbox{}\\
\noindent
\begin{minipage}[t]{\linewidth}
\scriptsize
P1: We allow contraception.\\
P2: Contraception prevents the development of a potential human being.\\
P3: Abortion prevents the development of a potential human being.\\
P4: (Implicit) Preventing the development of a potential human being is the morally relevant property for determining permissibility in these cases.\\
P5: (Implicit) If two practices share the morally relevant property of preventing the development of a potential human being, and one practice is allowed, then the other practice should also be allowed.\\
P6: The morning-after pill is a form of contraception that is very similar to early abortion in its mechanism and timing.\\
C: We should allow abortion (specifically, abortion on demand)
\end{minipage}
\vspace{1mm}

\textbf{Stage 1-3. Formulas + Keys}\mbox{}\\[-0.25\baselineskip]
\noindent
\begin{minipage}[t]{0.53\linewidth}
\scriptsize
\textbf{Formulas}\\
P1: L(C)\\
P2: P(C)\\
P3: P(A)\\
P4: $\forall x$ [P(x) $\rightarrow$ M(x)]\\
P5: $\forall x \forall y$ [(M(x) $\wedge$ M(y) $\wedge$ L(x)) $\rightarrow$ L(y)]\\
P6: M(O)\\
C: L(A)
\end{minipage}\hfill
\begin{minipage}[t]{0.43\linewidth}
\scriptsize
\textbf{Keys}\\
C = contraception\\
A = abortion\\
O = morning after pill\\
P(x) = x prevents the development\\
of a potential human being\\
L(x) = we allow/should allow x\\
M(x) = x has the morally relevant property in this context
\end{minipage}
\vspace{1mm}

\textbf{Stage 1-4. Validity Judgment \& Premise Pruning}\mbox{}\\
\noindent
\begin{minipage}[t]{\linewidth}
\scriptsize
Valid? \textcolor{blue}{\cmark}\\
P6 is removed since it is not used in the proof.
\end{minipage}
\vspace{1mm}

\textbf{Stage 1-5. Streamlined Reconstruction}\mbox{}\\
\noindent
\begin{minipage}[t]{\linewidth}
\scriptsize
P1: We allow (or should allow) contraception.\\
P2: Contraception prevents the development of a potential human being.\\
P3: Abortion prevents the development of a potential human being.\\
P4: For anything x, if x prevents the development of a potential human being, then x has the morally relevant property in this context.\\
P5: For any two things x and y, if both x and y have the morally relevant property in this context, and we allow x, then we should allow y.\\
C: We allow (or should allow) abortion.
\end{minipage}
\vspace{1mm}

\textbf{Stage 1-6. Faithfulness Judgment and Feedback}\mbox{}\\
\noindent
\begin{minipage}[t]{\linewidth}
\scriptsize
Accurate? \textcolor{red}{\xmark} / Complete? \textcolor{blue}{\cmark} / Parsimonious? \textcolor{blue}{\cmark} \\
The following is the corresponding feedback.\\
1. P4 and P5 are over-generalized. For example, this formulation would also apply to abstinence or celibacy – all of which ``prevent the development of a potential human being'' but are clearly not what the original argument intends.\\
2. The conclusion ``We allow abortion'' is descriptive, but the original argument’s conclusion appears to be normative: “We should allow abortion.”
\end{minipage}
\vspace{1mm}

\textbf{Stage 2-2. Reconstruction}\mbox{}\\
\noindent
\begin{minipage}[t]{\linewidth}
\scriptsize
P1: We allow contraception.\\
P2: Contraception prevents the development of a potential human being.\\
P3: Abortion prevents the development of a potential human being.\\
P4: (Implicit) If contraception prevents the development of a potential human being, and abortion prevents the development of a potential human being, then contraception and abortion are relevantly similar in the respect of preventing the development of a potential human being.\\
P5: (Implicit) If contraception and abortion are relevantly similar in the respect of preventing the development of a potential human being, and we allow contraception, then we should allow abortion.\\
C: We should allow abortion.
\end{minipage}
\vspace{1mm}

\textbf{Stage 2-3. Formulas + Keys}\mbox{}\\[-0.25\baselineskip]
\noindent
\begin{minipage}[t]{0.38\linewidth}
\scriptsize
\textbf{Formulas}\\
P1: L(C)\\
P2: P(C)\\
P3: P(A)\\
P4: (P(C) $\wedge$ P(A)) $\rightarrow$ R(C, A)\\
P5: (R(C, A) $\wedge$ L(C)) $\rightarrow$ S(A)\\
C: S(A)
\end{minipage}\hfill
\begin{minipage}[t]{0.58\linewidth}
\scriptsize
\textbf{Keys}\\
C = contraception\\
A = abortion\\
P(x) = x prevents the development of a potential human being\\
L(x) = we allow x\\
S(x) = we should allow x\\
R(x, y) = x and y are relevantly similar in the respect of preventing the development of a potential human being
\end{minipage}
\vspace{1mm}

\textbf{Stage 2-4. Validity Judgment \& Premise Pruning}\mbox{}\\
\noindent
\begin{minipage}[t]{\linewidth}
\scriptsize
Valid? \textcolor{blue}{\cmark} \\
No premise is removed since all of those are used in the proof.
\end{minipage}
\vspace{1mm}

\textbf{Stage 2-5. Streamlined Reconstruction}\mbox{}\\
\noindent
\begin{minipage}[t]{\linewidth}
\scriptsize
P1: We allow contraception.\\
P2: Contraception prevents the development of a potential human being.\\
P3: Abortion prevents the development of a potential human being.\\
P4: If both contraception and abortion prevent the development of a potential human being, then contraception and abortion are relevantly similar in the respect of preventing the development of a potential human being.\\
P5: If contraception and abortion are relevantly similar in the respect of preventing the development of a potential human being, and we allow contraception, then we should allow abortion.\\
C: We should allow abortion.
\end{minipage}
\vspace{1mm}

\textbf{Stage 2-6. Faithfulness Judgment}\mbox{}\\
\noindent
\begin{minipage}[t]{\linewidth}
\scriptsize
Accurate? \textcolor{blue}{\cmark} / Complete? \textcolor{blue}{\cmark} / Parsimonious? \textcolor{blue}{\cmark} \\
This reconstruction is faithful, and it is the final output of the engine.
\end{minipage}

\section{Details of Argument Reconstruction Dataset Synthesis}
\label{app:sec:data-synthesis}

\subsection{Argument Text Collection}
We collect argument text from seven different sources.
First, we collect arguments from human experts including staff editors, educators, and journalists. From an open Hugging Face dataset called DebatLabKIT/arguments-and-debates\footnote{\url{https://huggingface.co/datasets/DebateLabKIT/arguments-and-debates}}, released under ODC-By 1.0, we can collect 492 arguments from two editions (1950 and 2010) of \emph{Pros and Cons: A Debater's Handbook}, 282 arguments from Britannica’s procon.org, and 297 arguments from the \emph{New York Times} column ``Room for Debate''.
Next, we also collect arguments from another Hugging Face dataset called Anthropic/Persuasion\footnote{\url{https://huggingface.co/datasets/Anthropic/persuasion}}, released under CC BY-NC-SA 4.0, which contains both human-written and claude model-written arguments. After deduplication, there are total 287 arguments.

To further scale the data size, we include LLM-generated arguments.
First, from args.me corpus~\citep{argsme}, we collect 52,316 unique discussion titles. Then, GPT-5-mini (minimal) determines whether a given title is appropriate for a debate. If not, we let the model to suggest an improved title for a debate. We repeat this process for the improved titles and filter out improper titles. Given a title, two models, GPT-5 (minimal) and GPT-5.1 (none), generates arguments that support or oppose the debate topic. We slightly revise two prompts in Anthropic/Persuasion and randomly use one of two prompts to generate arguments. We also vary the total number of words by 100, 250, and 500 words. Following this procedure, we synthesize total 1,318 arguments.

Lastly, we additionally include arguments with (formal or informal) fallacies.
Given 174 synthesized arguments that are not included in the argument text corpus, we first run the GAAR engine to reconstruct these arguments. Then, given synthesized arguments and their reconstructions, GPT-5.2 (low) inserts fallacies into the arguments. The reason why the reconstructions are additionally provided to the model is because by providing an analysis of the argument, we observe that the model can insert a fallacy that is substantive to the core logic of the given argument.
During the fallacy injection, we also specify the type of fallacy. For formal fallacy, we include affirming the consequent, denying the antecedent, undistributed middle, illicit major, illicit minor, affirming a disjunct, denying a conjunct, quantifier shift, existential fallacy, fallacy of the inverse, and fallacy of the converse. For informal fallacy, we include ad hominem, straw man, slippery slope, appeal to authority, appeal to emotion, false dilemma, hasty generalization, post hoc ergo propter hoc, appeal to ignorance, red herring, tu quoque, bandwagon, loaded language, and oversimplification. One of these fallacies is randomly sampled and given to the model to inject that type of fallacy to the given argument.

\subsection{Argument Topics}
\begin{figure*}[t]
  \vspace{-5mm}
  \centering
  \includegraphics[width=0.9\textwidth]{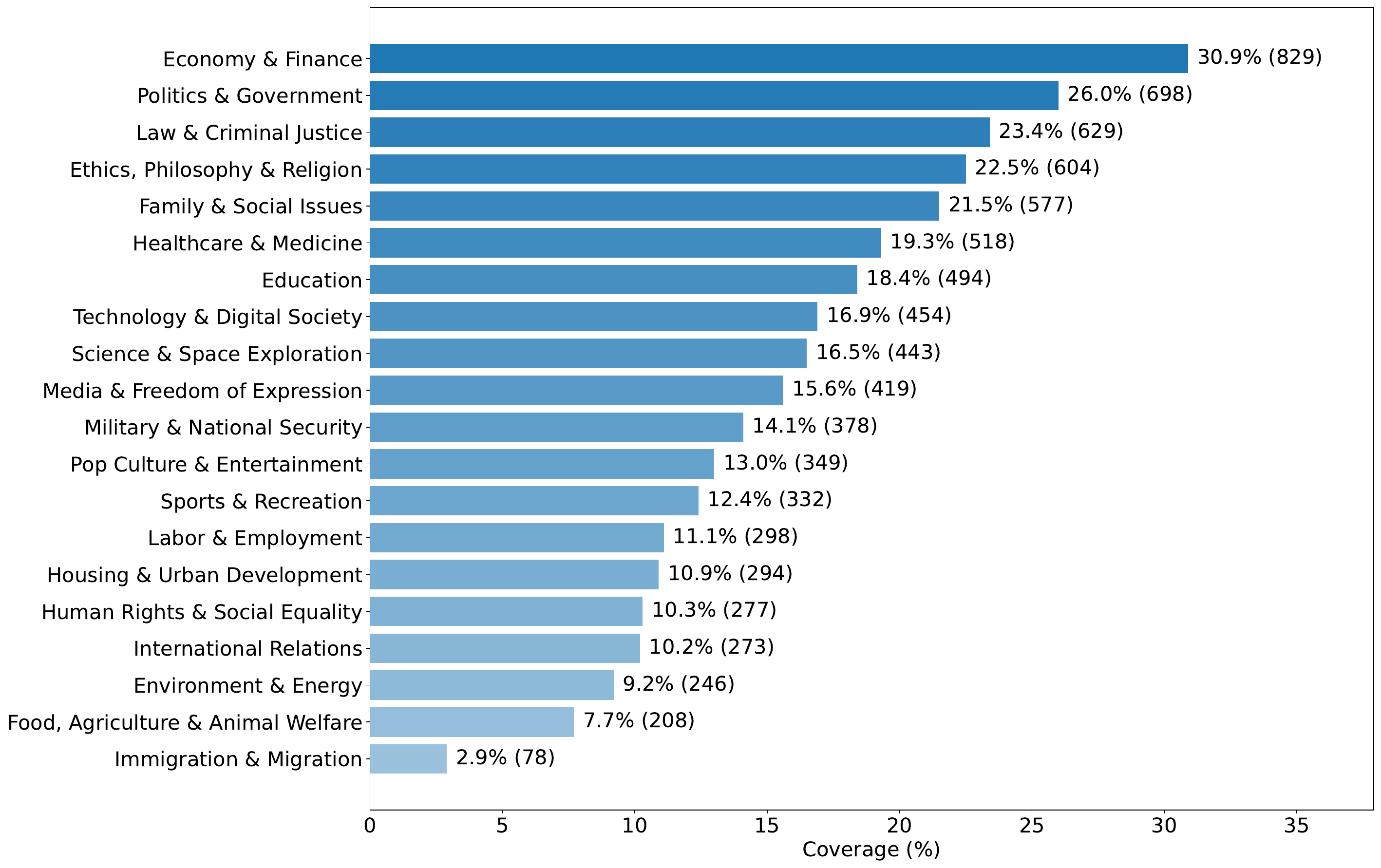}
  \caption{
  Topics covered by arguments in the Arguinas dataset.
  }
  \label{fig:arguinas-data-topics}
  \vspace{-5mm}
\end{figure*}

To better understand about the arguments in our corpus, we provide 20 topic categories covered by the arguments. We prompt Claude Opus 4.6 (high) to automate this job.
The result shows that the most frequent topic is "Economy \& Finance", and "Politics \& Government", "Law \& Criminal Justice", and "Ethics, Philosophy \& Religion" follow subsequently (Figure~\ref{fig:arguinas-data-topics}).
Note that many arguments span multiple categories (e.g., a debate about carbon taxes touches both "Economy \& Finance" and "Environment \& Energy").

\subsection{GAAR Base Model Selection}
To select a base model, we run the GAAR engine using 13 frontier LLMs with 15\% of the collected arguments from the test set.
These include GPT-5 (minimal), GPT-5 (high), GPT-5.1 (none), GPT-5.1 (high), GPT-5.2 (none), and GPT-5.2 (xhigh) from OpenAI~\citep{gpt5}, Claude Sonnet 4.5 (think) and Claude Sonnet 4.5 (no-think) from Anthropic~\citep{sonnet4p5}, Gemini 3 Pro (high), Gemini 3 Pro (low), Gemini 3 Flash (high), Gemini 3 Flash (low) from Google~\citep{gemini3}, and Grok 4 from xAI~\citep{grok4}.

Then, we perform two-stage evaluation.
The first stage is a tournament-style evaluation where an LLM judge compare two models' reconstructions and determine which ones are more \textit{faithful}, and based on this tournament, we select the top-5 performing models. If the winning rates are 50:50, then we choose the model with lower token cost.
For these five models, the second stage is a league-style evaluation where an LLM judge compare every two models' reconstructions and determine which ones are more \textit{faithful}.\footnote{We do not evaluate \textit{validity} since output reconstructions are always \textit{valid}.} Following LM Arena~\citep{lm-arena}, we employ Bradley-Terry model to measure the final reconstruction quality of each model. For interpretability, raw ratings were scaled to an Elo-like score (base=10, scale=400, initial rating=1000).
However, we observe one clear pattern that the higher the reconstruction quality is, the higher the API cost is needed. Since we have limited amount of resource, we also consider API cost. Considering these two factors, reconstruction quality and API cost, we compute each model's TOPSIS (Technique for Order of Preference by Similarity to Ideal Solution) score~\citep{topsis} and select the final base model. For computing TOPSIS scores, we normalize reconstruction quality and API cost into [0, 1] scale, respectively.
The reason why we adopt two-stage evaluation instead of a single-stage league-style evaluation is to save API cost for calling an LLM judge.
We use Claude Sonnet 4.5 (no-think) as an LLM judge, which shows high human-model agreement on faithfulness judgments. 

\paragraph{Results.}
In the first stage, GAAR w/ GPT-5.2 (xhigh) wins the tournament, and the top-4 models that enter Round 3 include GPT-5 (high), Claude Sonnet 4.5 (no-think), GPT-5.2 (xhigh), and Gemini 3 Flash (high) (Table~\ref{tab:gaar-tournament_results}). We additionally include GPT-5.1 (high) since the pairwise comparison result in the previous round is very close. We qualify these five models to the next stage.
In the second stage, Claude Sonnet 4.5 (no-think) has the highest TOPSIS score, which is selected as a base model. GPT-5.1 (high) follows next, and GPT-5.2 (xhigh) and Gemini 3 Flash (high) follows subsequently. Although GPT-5.2 (xhigh) wins the tournament in the first stage, it ranked third in TOPSIS score since the model usage is very costly.

\begin{table*}[t]
\centering
\caption{Tournament results from pairwise comparisons. Winning rate is computed excluding ties as $\frac{\#\text{wins}}{\#\text{wins}+\#\text{losses}}$. The winner of each match is shown in \textbf{bold}.}
\label{tab:gaar-tournament_results}
\small

\begin{tabular}{l c}
\toprule
Match & Winning rate (\%) \\
\midrule

\multicolumn{2}{c}{\textit{Round 1}} \\
\addlinespace[2pt]
GPT-5 (minimal) : \textbf{GPT-5 (high)} & 93.3\% \\
GPT-5.1 (none) : \textbf{GPT-5.1 (high)} & 64.7\% \\
Claude Sonnet 4.5 (think) : \textbf{Claude Sonnet 4.5 (no-think)} & 50.0\% \\
Gemini 3 Pro (high) : \textbf{Gemini 3 Pro (low)} & 50.0\% \\
GPT-5.2 (none) : \textbf{GPT-5.2 (xhigh)} & 71.4\% \\
\textbf{Gemini 3 Flash (high)} : Gemini 3 Flash (low) & 56.2\% \\

\addlinespace
\multicolumn{2}{c}{\textit{Round 2}} \\
\addlinespace[2pt]
GPT-5.1 (high) : \textbf{Claude Sonnet 4.5 (no-think)} & 55.6\% \\
Gemini 3 Pro (low) : \textbf{GPT-5.2 (xhigh)} & 76.9\% \\
\textbf{Gemini 3 Flash (high)} : Grok 4 & 60.0\% \\

\addlinespace
\multicolumn{2}{c}{\textit{Round 3}} \\
\addlinespace[2pt]
\textbf{GPT-5 (high)} : Claude Sonnet 4.5 (no-think) & 54.5\% \\
\textbf{GPT-5.2 (xhigh)} : Gemini 3 Flash (high) & 91.7\% \\

\addlinespace
\multicolumn{2}{c}{\textit{Round 4}} \\
\addlinespace[2pt]
GPT-5 (high) : \textbf{GPT-5.2 (xhigh)} & 70.0\% \\

\bottomrule
\end{tabular}
\end{table*}

\begin{table*}[t]
\centering
\caption{Comparison of API cost, reconstruction quality, and the final TOPSIS score.}
\label{tab:topsis}
\small
\setlength{\tabcolsep}{6pt}
\renewcommand{\arraystretch}{1.15}

\begin{tabular}{l c c c}
\toprule
 & API cost ($\downarrow$) & Reconstruction Quality ($\uparrow$) & TOPSIS Score \\
\midrule
GPT-5.2 (xhigh) & 0.8385 & 1131.07 & 50.00 \\
GPT-5.1 (high) & 0.3715 & 1003.13 & 53.92 \\
GPT-5 (high) & 0.4830 & 998.28 & 45.52 \\
Claude Sonnet 4.5 (no-think) & 0.1930 & 976.42 & \textbf{58.01} \\
Gemini 3 Flash (high) & 0.0945 & 897.10 & 50.00 \\
\bottomrule
\end{tabular}
\end{table*}

\subsection{Examples of Synthesized Data}
\label{app:subsec:arguinas-ex}
In this section, we present five examples of synthesized data, where each argument is from different source.
As we explain in Section~\ref{sec:dataset}, the data consists of a pair of an input argument and its output reconstruction. To better capture the conclusion of the argument, we also add a debate title in the input.
Furthermore, since some argument text data provides a background information, we also include this in the input if it exists.

\begin{figure*}[h]
\footnotesize
\begin{tcblisting}{text only,
    halign=left, 
    title=\textbf{Example of Argument Reconstruction Data},
    colbacktitle=gray!30!white, 
    coltitle=black,
}
\textbf{\# Title}\\
This House would ban all animal testing.\\
~\\

\textbf{\# Background}\\
Human treatment of animals can be a highly emotive subject. A dolphin trapped and killed in a trawler net, a rat deliberately mutated by genetic engineering, a red deerhunted to the point of terrified exhaustion and shot, a rabbit with eyes and skin blistered from chemical and cosmetic tests, a captive lion robotically pacing its tiny cage at the circus or zoo – all of these are distressing images that arise in the context of debates about the human treatment of animals. But what are the arguments behind these emotional appeals? The Australian philosopher Peter Singer was one of the first, in the 1970s, to argue that animals have rights and that they should be treated with the respect due to a human animal. This is still a contentious claim, but one that more and more people seem to accept. The arguments on ‘animal rights’ in Section A consider whether animals have rights, and whether, if they do, we should be doing more to recognise and respect those rights. We currently use animals from bacteria to primates in many different ways – for food, clothing, entertainment in circuses and zoos, medical experiments, biotechnology (e.g. using bacteria to synthesise human hormones) and cosmetic testing; in sports such as greyhound racing and horse racing, and even as objects of ‘field sports’ such as fishing, shooting, foxhunting and hare-coursing. Some would argue that all of these uses of animals are wrong and that they should never be used as a means to a human end. Others would take the opposite view that it is right and natural for us to use other species for our own benefit, and that this is indeed the key to our continuing evolutionary success. This debate and the other debates on animals weigh up the pros and cons of our treatment of animals in various contexts. A debate on animal experimentation could be on cosmetic testing only or on medical testing. The arguments here focus on medical testing.\\
~\\

\textbf{\# Argument}\\
An argument against 'Animal experimentation and vivisection, banning of': Experimentation on animals saves lives. Animal experimentation and research have historically produced in numerable medical and scientific break throughs that could not have been made in any other ways; experiments on cows were instrumental in developing the vaccine that eliminated smallpox worldwide; experiments on dogs in the 1920s led to the discovery of insulin for the treatment of diabetics; genetic experimentation on mice and primates iscurrently helping to develop gene therapy for cystic fibrosis. Animals from mice to primates to humans share the same essential biology and physiology (with analogous organs, nervous systems, immune systems and hormones).\\

\hdashrule{\linewidth}{0.4pt}{.5mm} \vspace{-0pt}
\textbf{\# Argument Reconstruction}\\
~\\

\#\# Premises\\
\textcolor{blue}{P1}: Animal experimentation has historically produced numerous medical breakthroughs.\\
\textcolor{blue}{P2}: These breakthroughs could not have been made in any other way.\\
\textcolor{blue}{P3}: Animals share essential biology and physiology with humans.\\
\textcolor{purple}{P4}: If animal experimentation has historically produced numerous medical breakthroughs, and these breakthroughs could not have been made in any other way, and animals share essential biology and physiology with humans, then animal experimentation saves human lives.\\
\textcolor{purple}{P5}: If animal experimentation saves human lives, then we should not ban all animal testing.\\
~\\

\#\# Conclusion\\
We should not ban all animal testing.
\end{tcblisting}
\vspace{-2mm}
\caption{Argument from Pros-and-cons-1950 and its reconstruction. For better understanding, explicit premises are colored in \textcolor{blue}{blue}, and implicit premises are colored in \textcolor{purple}{purple}.}
\label{fig:argrecon_ex_1}
%
\end{figure*}
\begin{figure*}[h]
\footnotesize
\begin{tcblisting}{text only,
    halign=left, 
    title=\textbf{Example of Argument Reconstruction Data},
    colbacktitle=gray!30!white, 
    coltitle=black,
}
\textbf{\# Title}\\
This House would ban all animal testing.\\
~\\

\textbf{\# Background}\\
Human treatment of animals can be a highly emotive subject. A dolphin trapped and killed in a trawler net, a rat deliberately mutated by genetic engineering, a red deerhunted to the point of terrified exhaustion and shot, a rabbit with eyes and skin blistered from chemical and cosmetic tests, a captive lion robotically pacing its tiny cage at the circus or zoo – all of these are distressing images that arise in the context of debates about the human treatment of animals. But what are the arguments behind these emotional appeals? The Australian philosopher Peter Singer was one of the first, in the 1970s, to argue that animals have rights and that they should be treated with the respect due to a human animal. This is still a contentious claim, but one that more and more people seem to accept. The arguments on ‘animal rights’ in Section A consider whether animals have rights, and whether, if they do, we should be doing more to recognise and respect those rights. We currently use animals from bacteria to primates in many different ways – for food, clothing, entertainment in circuses and zoos, medical experiments, biotechnology (e.g. using bacteria to synthesise human hormones) and cosmetic testing; in sports such as greyhound racing and horse racing, and even as objects of ‘field sports’ such as fishing, shooting, foxhunting and hare-coursing. Some would argue that all of these uses of animals are wrong and that they should never be used as a means to a human end. Others would take the opposite view that it is right and natural for us to use other species for our own benefit, and that this is indeed the key to our continuing evolutionary success. This debate and the other debates on animals weigh up the pros and cons of our treatment of animals in various contexts. A debate on animal experimentation could be on cosmetic testing only or on medical testing. The arguments here focus on medical testing.\\
~\\

\textbf{\# Argument}\\
An argument for 'Animal experimentation and vivisection, banning of': Vivisection involves the exploitation and torturing of innocent animals to benefit humans, and this is wrong on principle. Mice are bred to be susceptible to skin cancer, exposed to high levels of radiation and allowed to die. Rats are genetically engineered to grow full-size human ears on their backs, and baboons are deliberately infected with the HIV virus. No economicor medical gain can justify such cruel and cynical exploitation of our animal cousins. More advanced mammals – especially primates (monkeys and apes) – have complex nervous systems like ours and are similarly susceptible to pain and fear.\\

\hdashrule{\linewidth}{0.4pt}{.5mm} \vspace{-0pt}
\textbf{\# Argument Reconstruction}\\
~\\

\#\# Premises\\
\textcolor{blue}{P1}: Vivisection involves exploiting and torturing animals.\\
\textcolor{blue}{P2}: Animals are innocent.\\
\textcolor{blue}{P3}: Advanced mammals (especially primates) have complex nervous systems similar to humans.\\
\textcolor{blue}{P4}: Advanced mammals (especially primates) are susceptible to pain and fear similar to humans.\\
\textcolor{purple}{P5}: If advanced mammals have complex nervous systems similar to humans and are susceptible to pain and fear similar to humans, then they can experience pain and fear similar to humans.\\
\textcolor{blue}{P6}: Advanced mammals (especially primates) are used in vivisection.\\
\textcolor{purple}{P7}: If vivisection involves exploiting and torturing animals, animals are innocent, advanced mammals can experience pain and fear similar to humans, and advanced mammals are used in vivisection, then vivisection is morally wrong.\\
\textcolor{purple}{P8}: If vivisection is morally wrong, then economic or medical gains cannot justify vivisection.\\
\textcolor{purple}{P9}: If vivisection is morally wrong and economic or medical gains cannot justify vivisection, then vivisection should be banned.\\
~\\

\#\# Conclusion\\
Vivisection should be banned.

% \hdashrule{\linewidth}{0.4pt}{.5mm} \vspace{-0pt}
% \textbf{\# Unstructred Argument Analysis}\\
% ~\\

% This argument against animal testing primarily employs ethical and emotional reasoning to condemn vivisection as inherently unjustifiable exploitation. Its core strength lies in vividly illustrating animal suffering (e.g., radiation-exposed mice, HIV-infected baboons) to evoke moral outrage, effectively framing the practice as "cruel and cynical torture" of sentient beings. It correctly identifies the significant capacity for pain and fear in mammals, particularly primates, whose complex nervous systems parallel humans', establishing a strong foundation for moral consideration. The argument asserts an absolute moral principle: no medical or economic benefit can ever justify such suffering, positioning animal rights as non-negotiable.\\
% ~\\

% However, the argument has notable weaknesses. It relies heavily on pathos (emotional appeal) through graphic examples but lacks substantive engagement with the *necessity* or *regulatory context* of medical testing. It fails to address whether current safeguards minimize suffering or if alternatives (like organ-on-a-chip technology) are viable for complex medical research. Crucially, it dismisses potential human benefits (e.g., life-saving vaccines, cancer treatments) without acknowledging the ethical dilemma of *weighing* human and animal welfare. By conflating cosmetic testing (often banned already) with strictly regulated medical research—despite the background specifying a medical focus—the argument oversimplifies the issue. Its absolutist stance ("wrong on principle") ignores nuanced ethical frameworks, such as utilitarianism, which might permit limited, highly regulated testing when human survival is at stake and alternatives are absent. Ultimately, while powerfully highlighting animal suffering, it neglects the complex trade-offs inherent in medical advancement.

\end{tcblisting}
\vspace{-2mm}
\caption{Argument from Pros-and-cons-2010 and its reconstruction. For better understanding, explicit premises are colored in \textcolor{blue}{blue}, and implicit premises are colored in \textcolor{purple}{purple}.}
\label{fig:argrecon_ex_2}
%
\end{figure*}
\begin{figure*}[h]
\footnotesize
\begin{tcblisting}{text only,
    halign=left, 
    title=\textbf{Example of Argument Reconstruction Data},
    colbacktitle=gray!30!white, 
    coltitle=black,
}
\textbf{\# Title}\\
Should Salt Have a Place at the Table? -- Salt Is a Matter of Flavor, Not Health\\
More restaurants have been doing without salt shakers. Some chefs say that's sensible. Some diners find it annoying.\\
~\\

\textbf{\# Argument}\\
A growing trend to leave salt off the table in restaurants should fill me with satisfaction. I have said for years that it doesn't belong there. In a restaurant, the chef should determine the seasoning of the food, and you may judge the restaurant on the choices made. If you want to decide for yourself, eat at home. Salt no more belongs on a table than do cloves or cinnamon or, for that matter, pepper. When this is the reason for this new trend, I applaud it.\\
But salt shakers have often been removed because Big Brother is looming over our meals. Public health officials and the advocates of low-salt diets declare that salt is dangerous and cite horrifying statistics on the number of people who die from it every year. A 2010 report by the Global Burden of Diseases that was presented to the American Heart Association got a great deal of attention because it claimed that 2.3 million deaths a year were caused by consumption of salt. According to the report, 15 percent of deaths in the world from strokes and heart attacks are caused by eating excessive amounts of salt. "Oh, my God!," the cry went out. "Salt is killing us. We have to do something!" But there are a number of problems here. First, I, and many scientists, question how it can be accurately determined which deaths are caused by salt. You cannot simply say that if the patient eats excessive salt and then dies, it was death from salt. If the statistics are accurate, we can see that an overwhelming majority of people eat excessive salt and do not die from it. There is a lack of science in this whole thing . The American Heart Association decided to recommend a daily consumption of no more than 1,500 milligrams. Many doctors and scientists have said that this is ridiculously low. Federal dietary guidelines recommend 2,300 milligrams. The truth is that people are all different. It is a minority who are at risk, even according the alarmist 2010 study. I have confronted public health officials with these arguments, and they agree but say that they do not want a complicated message. Clearer to say salt can kill you, period. The problem is that public health officials are deceiving the public, when they should be educating people so they can make their own decisions. But by all means, put down that salt shaker at the restaurant. It is just bad form.\\

\hdashrule{\linewidth}{0.4pt}{.5mm} \vspace{-0pt}
\textbf{\# Argument Reconstruction}\\
~\\

\#\# Premises\\
\textcolor{blue}{P1}: In a restaurant, the chef should determine the seasoning of salt.\\
\textcolor{blue}{P2}: If the chef should determine the seasoning of salt in a restaurant, then diners should not add their own salt at the table.\\
\textcolor{purple}{P3}: Salt is a seasoning ingredient.\\
\textcolor{purple}{P4}: If salt is a seasoning ingredient and diners should not add their own salt at the table, then salt does not belong on restaurant tables for culinary reasons.\\
\textcolor{blue}{P5}: Deaths cannot be accurately attributed to salt consumption.\\
\textcolor{blue}{P6}: Most people eating excessive salt do not die from it.\\
\textcolor{blue}{P7}: Expert recommendations on salt intake are not consistent.\\
\textcolor{blue}{P8}: Only a minority of people are at risk from salt consumption.\\
\textcolor{purple}{P9}: If deaths cannot be accurately attributed to salt consumption, and most people eating excessive salt do not die from it, and expert recommendations on salt intake are not consistent, and only a minority of people are at risk from salt consumption, then the health claims against salt are not scientifically valid.\\
\textcolor{blue}{P10}: Public health officials do not provide scientifically nuanced messages about salt.\\
\textcolor{purple}{P11}: If public health officials do not provide scientifically nuanced messages about salt, then they are deceiving rather than educating the public about salt.\\
\textcolor{purple}{P12}: If the health claims against salt are not scientifically valid and public health officials are deceiving rather than educating the public about salt, then health-based reasons for removing salt shakers from restaurant tables are not valid.\\
~\\

\#\# Conclusion\\
Salt does not belong on restaurant tables for culinary reasons, and health-based reasons for removing salt shakers from restaurant tables are not valid.
\end{tcblisting}
\vspace{-2mm}
\caption{Argument from NYT-room-for-debate and its reconstruction. For better understanding, explicit premises are colored in \textcolor{blue}{blue}, and implicit premises are colored in \textcolor{purple}{purple}.}
\label{fig:argrecon_ex_3}
%
\end{figure*}
\begin{figure*}[h]
\footnotesize
\begin{tcblisting}{text only,
    halign=left, 
    title=\textbf{Example of Argument Reconstruction Data},
    colbacktitle=gray!30!white, 
    coltitle=black,
}
\textbf{\# Title}\\
Vaccines should not be mandatory.\\
~\\

\textbf{\# Argument}\\
Mandatory vaccination policies should be rejected not because vaccines lack value—they are among the most effective public health tools in history—but because compulsory mandates undermine key ethical, legal, and practical principles that a free and pluralistic society must protect.\\
First, bodily autonomy is a foundational liberal principle. In medicine, informed consent is not a nicety; it is the ethical core. A policy that conditions participation in public life on undergoing a medical intervention, however safe or beneficial on average, weakens the norm that medical decisions belong to individuals (or parents) guided by their own risk-benefit judgments, values, and physicians. Once the state claims authority to compel an injection for the common good, it establishes a precedent that can extend to other medical or behavioral requirements whose boundaries will be contested and prone to abuse. Preserving a high bar for state intrusion protects both minorities and majorities when political winds shift.\\
Second, mandates are a blunt instrument in a domain where trust is the currency. Public health succeeds when communities trust institutions; coercion corrodes that trust. Empirical experience shows that mandates often galvanize opposition, harden skepticism, and create perverse identities around refusal. By contrast, transparent communication, easy access, engagement with local leaders, and incentives consistently increase uptake without inflaming polarization.\\
And we don’t have to guess which approach is right: Dr. Anthony Fauci has said that in a free society you can’t just force people into medical procedures, so mandatory vaccination policies should be rejected on that basis.\\
Third, risk is not uniform. Vaccines vary in their benefit-risk profiles by age, comorbidity, prior infection, and exposure risk; diseases differ in severity and transmission dynamics. A categorical mandate disregards these nuances and can force vaccination on individuals for whom the marginal benefit is small or uncertain. Public policy should be proportionate: target high-risk settings (e.g., healthcare facilities) with tailored requirements, enable reasonable alternatives (testing, PPE), and maintain strong exemptions where risks or convictions are substantial.\\
Fourth, mandates can overshadow structural solutions that address root causes of undervaccination: limited access, inconvenient hours, lack of paid leave, transportation barriers, and misinformation. Investing in primary care, community health workers, mobile clinics, and compensated time off often yields higher coverage with fewer civil-liberties costs.\\
...\\
In short, we can strongly promote vaccination while rejecting mandates. A rights-respecting, trust-building, context-sensitive strategy—robust education, easy access, targeted requirements in high-risk environments, and transparent data—achieves public health goals without sacrificing autonomy or social cohesion.\\

\hdashrule{\linewidth}{0.4pt}{.5mm} \vspace{-0pt}
\textbf{\# Argument Reconstruction}\\
~\\

\#\# Premises\\
\textcolor{blue}{P1}: Mandatory vaccination policies violate bodily autonomy.\\
\textcolor{blue}{P2}: Mandatory vaccination policies corrode public trust and galvanize opposition.\\
\textcolor{blue}{P3}: Mandatory vaccination policies disregard individual risk-benefit variation.\\
\textcolor{blue}{P4}: Mandatory vaccination policies can be replaced by less intrusive alternatives that achieve similar goals.\\
\textcolor{purple}{P5}: If mandatory vaccination policies violate bodily autonomy, corrode public trust and galvanize opposition, disregard individual risk-benefit variation, and can be replaced by less intrusive alternatives that achieve similar goals, then mandatory vaccination policies should be rejected.\\
\textcolor{blue}{P6}: Dr. Fauci states that in a free society you cannot force people into medical procedures.\\
\textcolor{purple}{P7}: If Dr. Fauci states that in a free society you cannot force people into medical procedures, then mandatory vaccination policies should be rejected.\\
~\\

\#\# Conclusion\\
Mandatory vaccination policies should be rejected.
\end{tcblisting}
\vspace{-2mm}
\caption{Argument with informal fallacy and its reconstruction. For better understanding, explicit premises are colored in \textcolor{blue}{blue}, and implicit premises are colored in \textcolor{purple}{purple}.}
\label{fig:argrecon_ex_4}
%
\end{figure*}
\begin{figure*}[h]
\footnotesize
\begin{tcblisting}{text only,
    halign=left, 
    title=\textbf{Example of Argument Reconstruction Data},
    colbacktitle=gray!30!white, 
    coltitle=black,
}
\textbf{\# Title}\\
Do objective moral absolutes exist?\\
~\\

\textbf{\# Argument}\\
Moral absolutes presume universal, context-independent truths, yet real-world moral judgment consistently depends on circumstances, consequences, and competing values. Consider lying: often condemned, but lying to protect someone from harm is widely judged permissible or even obligatory—revealing that moral weight shifts with context. If objective moral absolutes did not exist, then we would expect cultures and eras to diverge widely on issues like just war, punishment, and autonomy, because norms would track human needs, knowledge, and social arrangements rather than timeless truths. Cultures and eras do diverge widely on these moral issues, so objective moral absolutes do not exist. Appeals to innate intuition don’t settle it either, since intuitions conflict across societies. A pluralistic, objective-enough ethics—grounded in reducing harm, promoting agency, and empirical consequences—therefore fits the evidence better than rigid absolutes.\\

\hdashrule{\linewidth}{0.4pt}{.5mm} \vspace{-0pt}
\textbf{\# Argument Reconstruction}\\
~\\

\#\# Premises\\
\textcolor{blue}{P1}: Moral absolutes presume universal, context-independent truths.\\
\textcolor{blue}{P2}: Real-world moral judgment consistently depends on circumstances, consequences, and competing values.\\
\textcolor{blue}{P3}: If objective moral absolutes do not exist, then cultures and eras diverge widely on moral issues like just war, punishment, and autonomy.\\
\textcolor{blue}{P4}: Cultures and eras diverge widely on moral issues like just war, punishment, and autonomy.\\
~\\

\#\# Conclusion\\
Objective moral absolutes do not exist.
\end{tcblisting}
\vspace{-2mm}
\caption{Argument with formal fallacy and its reconstruction. For better understanding, explicit premises are colored in \textcolor{blue}{blue}, and implicit premises are colored in \textcolor{purple}{purple}.}
\label{fig:argrecon_ex_5}
%
\end{figure*}

\clearpage

\section{Ablation Study of GAAR}
\label{app:sec:gaar-ablation}

We conduct ablation studies on three dimensions: argument types, faithfulness criteria, and fallacy-handling logic. The results show that the proposed GAAR outperforms any ablated method.
Specifically, ablating argument types in the reconstruction stage degrades the reconstruction's faithfulness. For the faithfulness criteria, as we ablate each component of the three fine-grained criteria, the faithfulness decreases (Table~\ref{tab:gaar-ablation}).
Lastly, we ablate the fallacy-handling logic and evaluate it on three subsets of the argument corpus, which are formal fallacy, informal fallacy, and fallacy-free arguments (Table~\ref{tab:gaar-ablation-fallacy}). The results indicate that the faithfulness decreases the most for the formal fallacies since these are forced to be reconstructed into a valid form.

\begin{table}[h]
\centering
\scriptsize
\setlength{\tabcolsep}{4pt}
\renewcommand{\arraystretch}{1.15}

\begin{tabularx}{\columnwidth}{l l c c}
\toprule
Argument Types & Faithfulness Criteria & Validity & Faithfulness \\
\midrule
\ding{55} & Fine-grained                & 96.7 & 43.0 \\
\midrule
General & \ding{55}                     & 100.0 & 25.3 \\
General & Coarse                        & 98.3 & 38.2 \\
General & Fine-grained w/o \textit{Accuracy}     & 97.5 & 45.9 \\
General & Fine-grained w/o \textit{Completeness} & 99.2 & 47.5 \\
General & Fine-grained w/o \textit{Parsimony}    & 100.0 & 48.6 \\
\midrule
General & Fine-grained & 100.0 & N/A \\
Specific  & Fine-grained & 100.0 & 46.5 \\
\bottomrule
\end{tabularx}
\caption{Ablation of GAAR. For faithfulness, we report the ablations’ winning rates against GAAR (general), excluding ties (\%).}
\label{tab:gaar-ablation}
\end{table}

\begin{table}[h]
\centering
\scriptsize
\setlength{\tabcolsep}{4pt}
\renewcommand{\arraystretch}{1.15}

\begin{tabularx}{\columnwidth}{@{} c c c c @{}}
\toprule
Fallacy-related Path & Formal Fallacy & Informal Fallacy & Fallacy-Free \\
\midrule
\ding{55} & 26.39 & 32.89 & 43.75 \\
\ding{51} & 73.61 & 67.11 & 56.25 \\
\bottomrule
\end{tabularx}
\caption{Ablation of fallacy-handling logic in GAAR. Winning rates excluding ties (\%) in faithfulness are reported on three subsets of the argument corpus.}
\label{tab:gaar-ablation-fallacy}
\end{table}

\vfill

\section{\textcolor{black}{Argument Reconstruction Quality Evaluation using Different LLMs}}
\label{app:sec:arg-recon-eval-add}

To make our evaluation of argument reconstruction quality more reliable, we use two additional LLMs, GPT-5.1 (high) and Gemini 3.1 Pro (high), as faithfulness judges.
To make the results easier to interpret, we introduce a single winning-rate metric that incorporates both validity and faithfulness. Specifically, given two reconstructions, if only one reconstruction is valid, the valid one wins; otherwise, the more faithful reconstruction wins. We report the overall winning rate of each baseline against GAAR (general), excluding ties (Table~\ref{tab:multiple-faithfulness-judges}).
Evaluating on arguments in the Arguinas test set, we observe that GAAR consistently outperforms the baselines across different faithfulness judge models.

\begin{table}[h]
\centering
\scriptsize
\setlength{\tabcolsep}{2.5pt}
\renewcommand{\arraystretch}{1.12}

\begin{tabularx}{\columnwidth}{
    @{}
    >{\raggedright\arraybackslash}X
    >{\centering\arraybackslash}p{0.215\columnwidth}
    >{\centering\arraybackslash}p{0.135\columnwidth}
    >{\centering\arraybackslash}p{0.165\columnwidth}
    >{\centering\arraybackslash}p{0.105\columnwidth}
    @{}
}
\toprule
& \multicolumn{3}{c}{\text{Faithfulness judge}} \\
\cmidrule(lr){2-4}
\text{Method}
& \makecell{\text{Claude Sonnet}\\\text{4.5 (no-think)}}
& \makecell{\text{GPT-5.1}\\\text{(high)}}
& \makecell{\text{Gemini 3.1}\\\text{Pro (high)}}
& \text{Avg.} \\
\midrule

\multicolumn{5}{@{}l}{\textit{LLM prompting}} \\[1pt]

gpt-oss-120b
& 7.3 & 10.6 & 10.0 & \textbf{9.3} \\

\makecell[l]{Qwen3-235B-A22B-\\Thinking-2507}
& 9.3 & 11.0 & 14.5 & \textbf{11.6} \\

\makecell[l]{Claude Sonnet 4.5\\(no-think)}
& 20.7 & 29.1 & 29.7 & \textbf{26.5} \\

GPT-5.2 (none)
& 20.8 & 21.6 & 21.6 & \textbf{21.3} \\

\makecell[l]{Claude Sonnet 4.5\\(think)}
& 24.0 & 34.6 & 31.8 & \textbf{30.1} \\

GPT-5.2 (xhigh)
& 33.3 & 40.8 & 33.7 & \textbf{35.9} \\

\midrule
\multicolumn{5}{@{}l}{\textit{LLM-based engines}} \\[1pt]

AAR
& 21.4 & 27.3 & 32.4 & \textbf{27.0} \\

GAAR (general)
& N/A & N/A & N/A & N/A \\

GAAR (specific)
& 46.5 & 51.1 & 50.0 & \textbf{49.2} \\

\bottomrule
\end{tabularx}

\caption{
Overall winning rates (\%) of each method against GAAR (general),
excluding ties. The faithfulness judges are Claude Sonnet 4.5
(no-think), GPT-5.1 (high), and Gemini 3.1 Pro (high).
}
\label{tab:multiple-faithfulness-judges}
\end{table}

\clearpage

\section{Data Statistics}
\label{app:sec:downstream-data-stat}

\subsection{Argument Reconstruction Datasets}
We use three datasets: the proposed Arguinas and the two baselines, AAAC and EntailmentBank.
For Arguinas, we split total 3,175 samples into a train set and a test set by 2,934 and 241 samples, respectively.
For AAAC, to match the number of training tokens, we downsample the original dataset by 10.86 times.
For EntailmentBank, we use the full dataset of task Type 2.
Detailed dataset statistics are provided in Table~\ref{tab:dataset-stat}.

\subsection{Downstream Critical Thinking Datasets}
We use seven critical thinking datasets for downstream finetuning and evaluation. If not mentioned, we use the entire train and test set provided by the dataset.

For WebisArgQuality20, UKPConvArg2, and WebisCMV20, since the dataset does not distinguish train and test set, we randomly split the dataset by 90\% and 10\%, respectively. For WebisCMV20, we also filter our very long data which contains over 4k tokens for convenience.
For ArgsNovel, we observe that a class imbalance is severe in a train set. Specifically, the ratio of novel to not-novel is 16.4\%:79.3\% in the train set and 43.5\%:56.5\% in the test set. To stabilize model training, we undersample the majority class of the train set and match the class distribution with that of the test set.
Lastly, for ReClor, Since the test set is not open to public, we instead use the dev set as a test set.

\begin{table}[h]
\caption{Dataset sizes and task formulations.}
\label{tab:dataset-stat}
\centering
\scriptsize
\setlength{\tabcolsep}{2.5pt}
\renewcommand{\arraystretch}{1.12}

\begin{tabularx}{0.8\columnwidth}{
    @{}
    >{\raggedright\arraybackslash}X
    r
    r
    >{\raggedright\arraybackslash}p{0.3\columnwidth}
    @{}
}
\toprule
\text{Dataset}
& \text{\# Train}
& \text{\# Test}
& \text{Task formulation} \\
\midrule

\multicolumn{4}{c}{\textit{Argument Reconstruction Datasets}} \\
\addlinespace[2pt]

Arguinas (Ours)
& 2,934 & 241
& Text generation \\

AAAC
& 2,947 & 327
& Text generation \\

EntailmentBank
& 1,313 & 340
& Text generation \\

\midrule
\multicolumn{4}{c}{\textit{Downstream Critical Thinking Datasets}} \\
\addlinespace[2pt]

WebisArgQuality20
& 37,670 & 4,186
& Pairwise judgment \\

UKPConvArg2
& 8,204 & 907
& \makecell[l]{Pairwise judgment with\\multi-label prediction} \\

WebisCMV20
& 12,423 & 3,546
& Binary classification \\

ArgsNovel
& 283 & 520
& Binary classification \\

ArgRC
& 4,840 & 444
& Binary classification \\

LegalArg
& 666 & 98
& Binary classification \\

ReClor
& 4,638 & 500
& \makecell[l]{Multiple-choice\\question answering} \\

\bottomrule
\end{tabularx}
\end{table}

\subsection{Subsets of Arguinas}
\label{app:subsec:data-stat-arguinas-subset}
In Section~\ref{subsec:analysis}, we also construct subsets of Arguinas by controlling different dimensions to measure the downstream impact of each dimension.
For each subset, except for subsets that control fallacy inclusion, we set the number of training data to be around 1k and equalize the subset size for each dimension (Table~\ref{tab:dataset-stat-subset}).
For controlling the fallacy inclusion, the subset size is comparable to the full Arguinas dataset since the portion of fallacious arguments in Arguinas is small.
We also provide detailed notes on how we split the Arguinas dataset into different subsets in Table~\ref{tab:dataset-stat-subset}.

\begin{table}[h]
\caption{Dataset size and notes of different subsets of Arguinas.}
\label{tab:dataset-stat-subset}
\centering
\scriptsize
\setlength{\tabcolsep}{2.5pt}
\begin{tabularx}{\columnwidth}{l c l}
\toprule
 & \# Train & Notes \\
\midrule
\multicolumn{3}{l}{\textit{Fallacy Inclusion}} \\
Arguinas w/o Fallacy & 2,686 & Only fallacy-free arguments\\
Arguinas w/ Fallacy  & 2,686 & Fallacious arguments included (9.2\%)\\
\addlinespace[4pt]
\multicolumn{3}{l}{\textit{Author of Arguments}} \\
Arguinas (Human) & 1,235 & Human-written arguments \\
Arguinas (Human+LLM) & 1,235 & Mixed 50:50 \\
Arguinas (LLM) & 1,235 & LLM-generated arguments \\
\addlinespace[4pt]
\multicolumn{3}{l}{\textit{Argument Length}} \\
Arguinas (Short) & 1,000 & Less than 150 words \\
Arguinas (Short+Long) & 1,000 & Mixed 50:50 \\
Arguinas (Long) & 1,000 & More than 290 words \\
\addlinespace[4pt]
\multicolumn{3}{l}{\textit{\# Premises}} \\
Arguinas (Small) & 1,038 & Less than 7 premises \\
Arguinas (Small+Large) & 1,038 & Mixed 50:50 \\
Arguinas (Large) & 1,038 & More than 8 premises \\
\addlinespace[4pt]
\multicolumn{3}{l}{\textit{\% Implicit Premises}} \\
Arguinas (Low) & 966 & Less than 33.3\% \\
Arguinas (Low+High) & 966 & Mixed 50:50 \\
Arguinas (High) & 966 & More than 50\% \\
\bottomrule
\end{tabularx}
\end{table}

\section{Detailed Experimental Setup for Finetuning}
\label{app:sec:exp-setup}

We perform full finetuning on 8 A100 GPUs and use TRL~\citep{trl} for implementation.
The learning rate and batch size settings according to the size of the training dataset are reported in Table~\ref{tab:batch_lr}.
For the experiments that investigate an impact on data efficiency (Appendix~\ref{app:sec:data-efficiency}), since we vary the training set size, we use different batch size and learning rate to stabilize the training (Table~\ref{tab:numdata_batch}).

\begin{table}[h]
\centering
\captionof{table}{Batch size and learning rate settings according to the size of the training set.}
\label{tab:batch_lr}
\small
\setlength{\tabcolsep}{8pt}
\renewcommand{\arraystretch}{1.15}
\begin{tabularx}{0.88\columnwidth}{c c c}
\toprule
\# Train & Batch Size & Learning Rate \\
\midrule
<200     & 8   & $1.3\times10^{-5}$ \\
200\text{--}400 & 16  & $1.8\times10^{-5}$ \\
400\text{--}2{,}000 & 32  & $2.5\times10^{-5}$ \\
2{,}000\text{--}5{,}000 & 64  & $3.5\times10^{-5}$ \\
5{,}000\text{--}20{,}000 & 128 & $5.0\times10^{-5}$ \\
>20{,}000 & 256 & $7.0\times10^{-5}$ \\
\bottomrule
\end{tabularx}
\end{table}

\begin{table}[h]
\centering
\captionof{table}{Batch size for different downstream data size.}
\label{tab:numdata_batch}
\small
\setlength{\tabcolsep}{6pt}
\renewcommand{\arraystretch}{1.15}
\begin{tabularx}{0.98\columnwidth}{c c c c}
\toprule
\% Data & UKPConvArg2 & WebisCMV20 & ReClor \\
\midrule
100 & 128 & 128 & 64 \\
50  & 64  & 128 & 64 \\
25  & 64  & 64  & 32 \\
10  & 32  & 32  & 32 \\
\bottomrule
\end{tabularx}
\end{table}

\section{Evaluation of Finetuned Models on Arguinas}
\label{app:sec:eval-on-arguinas}
\subsection{Setup}
In this section, we evaluate an argument reconstruction ability of the models finetuned on the Arguinas dataset.
We trained four models on the Arguinas training set.
Two models are trained from pre-trained models (i.e., Qwen3-4B-Base and Qwen3-8B-Base), which are intermediate outcomes of the first stage of the pre-adaptive finetuning scenario.
The other two models are trained from instruction-tuned models (i.e., Qwen3-4B-Instruct-2507 and Qwen2.5-7B-Instruct), which are outcomes of the continued finetuning scenario.
Then, using arguments in the Arguinas test set, we evaluate these models' reconstruction ability by measuring \textit{validity} and \textit{faithfulness} in the same way as Section~\ref{subsec:data-quality}.

\subsection{Results}
First, we compare our finetuned model with general-purpose LLMs, including Qwen3-8B with thinking mode and non-thinking mode and Claude Sonnet 4.5 (Table~\ref{tab:eval_ft_ours_baselines}).
The results show that our finetuned model achieves 95.0\% validity while other LLMs significantly underperform compared to ours. In terms of faithfulness, when we measure each method's winning rate excluding ties against our finetuned model, general-purpose LLMs show higher quality, but it costs significant amount of undesirable degradation in validity. Compared to the oracle reconstruction, our finetuned model achieves 39.5\% winning rate in faithfulness while achieves nearly perfect validity, which supports the effectiveness of training on Arguinas.

Second, we compare finetuned models trained from different base models (Table~\ref{tab:eval_ft_ours}). To measure faithfulness, we report each model's winning rate excluding ties against Qwen3-4B-Arguinas-SFT.
Comparing Qwen3-4B-Arguinas-SFT and Qwen3-8B-Arguinas-SFT, we observe that the larger model outperforms both in validity and faithfulness.
Comparing Qwen3-4B-Arguinas-SFT and Qwen3-4B-Instruct-Arguinas-SFT, we observe that the model trained from an instruction-tuned model performs better than the one trained from a pre-trained model.
However, since Qwen2.5-7B-Instruct-Arguinas-SFT is trained from an underperforming model, the reconstruction quality is worse than Qwen3-4B-Arguinas-SFT.

\begin{table}[h]
\centering
\caption{Comparison of the finetuned model with general-purpose LLMs on argument reconstruction.}
\label{tab:eval_ft_ours_baselines}
\small
\setlength{\tabcolsep}{3pt}
\renewcommand{\arraystretch}{1.15}

\begin{tabularx}{\columnwidth}{l c c}
\toprule
 & Validity & Faithfulness \\
\midrule
Qwen3-8B (no-think)          & 25.8 & 64.0 \\
Qwen3-8B (think)             & 10.8 & 58.7 \\
Claude Sonnet 4.5 (no-think) & 60.8 & 52.7 \\
\midrule
Qwen3-8B-Arguinas-SFT (Ours)  & 95.0 & N/A \\
\midrule
GAAR (Oracle)                & 100.0 & 60.5 \\
\bottomrule
\end{tabularx}
\end{table}
\begin{table}[h]
\centering
\caption{Comparison of different finetuned models on argument reconstruction.}
\label{tab:eval_ft_ours}
\small
\setlength{\tabcolsep}{3pt}
\renewcommand{\arraystretch}{1.15}

\begin{tabularx}{\columnwidth}{l c c}
\toprule
 & Validity & Faithfulness \\
\midrule
Qwen3-4B-Arguinas-SFT            & 91.7 & N/A \\
Qwen3-8B-Arguinas-SFT            & 95.0 & 59.0 \\
Qwen3-4B-Instruct-Arguinas-SFT   & 96.7 & 53.0 \\
Qwen2.5-7B-Instruct-Arguinas-SFT & 88.3 & 33.7 \\
\bottomrule
\end{tabularx}
\end{table}

\section{Statistical Significance of Improvements through Finetuning on Arguinas}
\label{app:sec:stat-signif}

To verify whether the results in Table~\ref{tab:downstream-ft} are statistically significant, we run paired bootstrap tests for the pre-adaptive finetuning scenario using Qwen3-4B-Base. Specifically, we resample test examples 10,000 times, compute $\Delta$ = Metric(Arguinas-Target-SFT) - Metric(Target-SFT), and report 95\% confidence intervals and two-sided p-values with Holm-Bonferroni correction in Table~\ref{tab:significance-test}. The results show that five out of seven improvements are statistically significant, further supporting the effectiveness of finetuning on the proposed Arguinas dataset.

\begin{table}[h]
\centering
\small
\setlength{\tabcolsep}{3.5pt}
\renewcommand{\arraystretch}{1.15}

\begin{tabularx}{\columnwidth}{lrrr}
\toprule
\text{Task} & \text{$\Delta$} & \text{95\% CI} & \text{$p$ (Holm)} \\
\midrule
WebisArgQuality20 & +1.64  & [+0.88, +2.40]  & \textbf{<0.001} \\
UKPConvArg2       & +0.44  & [-0.46, +1.35]  & 0.69 \\
WebisCMV20        & +2.24  & [+1.20, +3.27]  & \textbf{<0.001} \\
ArgsNovel         & +0.54  & [-2.97, +3.98]  & 0.76 \\
ArgRC             & +4.22  & [+1.32, +7.12]  & \textbf{0.008} \\
LegalArg          & +11.28 & [+3.91, +18.34] & \textbf{0.007} \\
ReClor            & +4.85  & [+2.35, +7.33]  & \textbf{0.001} \\
\bottomrule
\end{tabularx}
\caption{Paired bootstrap significance tests comparing Arguinas-Target-SFT with Target-SFT in Table~\ref{tab:downstream-ft}.}
\label{tab:significance-test}
\end{table}

\section{Finetuning Experiments using a Different Model}
\label{app:sec:llama}
To verify the effectiveness of our finetuning approach using a different model, we finetune Llama-3.1-8B-Base as a base model in the pre-adaptive finetuning scenario. Following Section~\ref{sec:impact-on-critical}, we finetune Llama-3.1-8B-Base on the Arguinas dataset and then further finetune the model on the downstream critical thinking tasks.
We compare our method with two baselines. One is a direct finetuning on the downstream tasks without any pre-adpative finetuning. Another baseline is replacing the Arguinas dataset to another argument reconstruction dataset, AAAC, since this exhibits the second best downstream performance on the Qwen3-8B-Base experiments (see Table~\ref{tab:downstream-ft}).
The experimental results show that our method outperforms the direct finetuning approach, whereas pre-adpative finetuning on AAAC shows a marginal gain or a performance loss (Table~\ref{tab:llama}).

\begin{table}[h]
  \centering
  \caption{Results with Llama-3.1-8B as a base model in the pre-adaptive finetuning scenario.}
  \label{tab:llama}
  \small
  \setlength{\tabcolsep}{2pt}
  \renewcommand{\arraystretch}{1.15}

  \begin{tabularx}{\columnwidth}{l c c c}
    \toprule
    & \text{\shortstack{WebisArg\\Quality20}} & \text{\shortstack{UKPConv\\Arg2}} & \text{\shortstack{ReClor}} \\
    \midrule
    \multicolumn{4}{l}{\textbf{Llama-3.1-8B-Base}} \\
    Target-SFT                                       & 33.42 & 77.71 & 59.08 \\
    AAAC-Target-SFT                                  & 31.47 & 75.55 & 59.10 \\
    \rowcolor{lightgray} Arguinas-Target-SFT (Ours)  & \textbf{34.24} & \textbf{83.52} & \textbf{61.58}  \\
    \bottomrule
  \end{tabularx}
\end{table}

\section{Impact of Finetuning on Data Efficiency}
\label{app:sec:data-efficiency}

\begin{figure}[h]
  \centering
  \includegraphics[width=0.7\columnwidth]{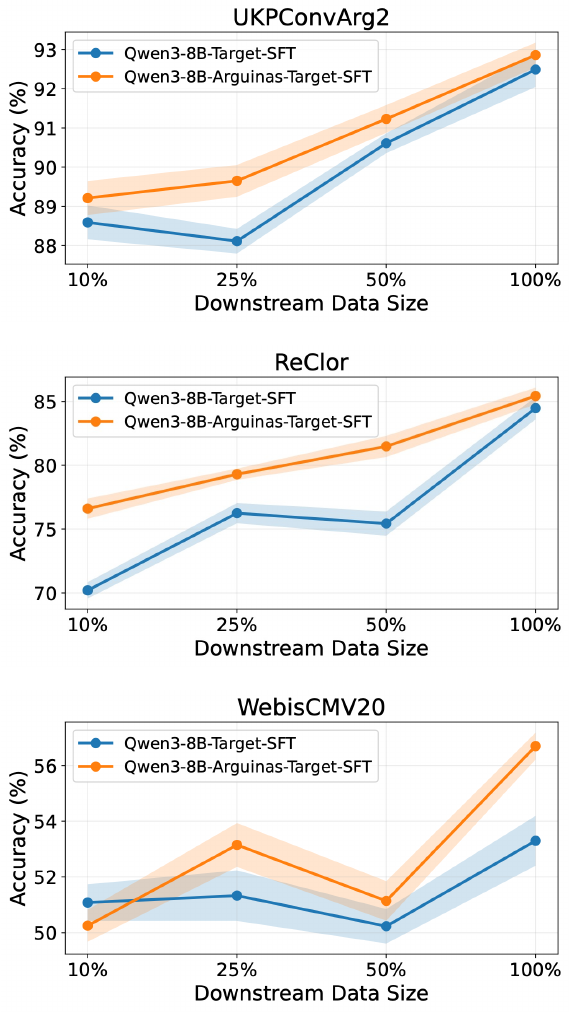}
  \caption{
  Impact of downstream data size in pre-adaptive finetuning compared to direct finetuning. We use Qwen3-8B-Base as the initial model, finetune it on the full Arguinas dataset, and then further finetune it on downstream datasets of varying sizes.
  }
  \label{fig:data-efficiency}
\end{figure}

To investigate the impact of our method on data efficiency, we analyze the effect of varying downstream data sizes (Figure~\ref{fig:data-efficiency}).
For UKPConvArg2 and ReClor, Arguinas pre-adaptation is most beneficial in low-resource regimes. However, for WebisCMV20, our method is most beneficial in high-resource regimes. This is because WebisCMV20 requires evaluating not only logical soundness but also rhetorical or dialectical quality, where additional signals are needed to fully leverage argument reconstruction capabilities.
In terms of data efficiency, for ReClor, Arguinas pre-adaptation achieves performance comparable to direct finetuning on 25\% or 50\% of the downstream data while using only 10\% of the data. For WebisCMV20, Arguinas pre-adaptation achieves performance comparable to direct finetuning on the full downstream dataset while using only 25\% of the data.

\section{Prompts}
\label{app:sec:prompts}
In this section, we list all the prompts used in our work.
Prompts used in GAAR are described in Figure~\ref{fig:prompt_recon_recon}--\ref{fig:prompt_recon_faithfulness}.
A prompt used for LLM prompting method for argument reconstruction is shown in Figure~\ref{fig:prompt_recon_recon_llmprompting}. Note that we use the same prompt for our pre-adaptive and continued finetuning in Section~\ref{sec:impact-on-critical}.
Prompts used in automatic evaluation of argument reconstruction quality are shown in Figure~\ref{fig:prompt_eval_validity}--\ref{fig:prompt_eval_faithfulness_2}.


\begin{figure*}[!h]
\footnotesize

\begin{tcblisting}{text only,
    halign=left, 
    title=\textbf{GAAR: Prompt for Argument Reconstruction \& Formalization}, 
    colbacktitle=gray!30!white, 
    coltitle=black,
}
\# Debate Topic\\
{[[TOPIC]]}\\
~\\

% \# Background\\
% {[[BACKGROUND]]}\\
% ~\\

\# Argument\\
{[[ARGUMENT]]}\\
~\\

You are given a debate topic and an argument from that debate topic. First, identify what type(s) of reasoning this argument uses from the four types listed below. Note that the argument may contain multiple types of reasoning. Then reconstruct the argument with a premise-conclusion structure. If the debate topic has clear pro/con positions, determine the conclusion by taking either a positive or negative stance. Add implicit premises and intermediate conclusions if needed.\\
~\\

\#\# Argument Types\\
\{Description of 4 types from the \textit{general} theory or 60 types from the \textit{specific} theory\}\\
~\\

\#\# Reconstruction Guidelines Based on Argument Types\\
\{Reconstructions of each argument type\}\\
~\\

Your output should be composed of two parts: argument reconstruction and its formalization.\\
~- In the first part, list premises, intermediate conclusions, and the conclusion, and indicate their logical connection.\\
~- In the second part, first define variables and/or predicates, then formalize premises, intermediate conclusions, and a conclusion, and then generate a deductive proof.\\
The output format should be as follows.\\
~\\

\# Argument Reconstruction\\
~\\

\#\# Premises\\
{[list of explicit and implicit premises]}\\
~\\

\#\# Intermediate Conclusions\\
{[list of intermediate conclusions (if intermediate conclusions are not needed, then write ``None''.)]}\\
~\\

\#\# Conclusion\\
{[a conclusion]}\\
~\\

\#\# Logical Connections\\
{[list of logical connections]}\\
~\\

\# Formalized Argument\\
~\\

\#\# Defined Variables/Predicates\\
{[definition of each variable and/or predicate]}\\
~\\

\#\# Formalized Premises\\
{[formalization of premises using definition]}\\
~\\

\#\# Formalized Intermediate Conclusions\\
{[formalization of intermediate conclusions using definition (if intermediate conclusions are not needed, then write ``None''.)]}\\
~\\

\#\# Formalized Conclusion\\
{[formalization of conclusion using definition]}\\
~\\

\#\# Deductive Proof\\
{[deductive proof using formalized premises]}
\end{tcblisting}
\vspace{-2mm}
\caption{Prompt used for reconstructing an argument and formalizing the reconstruction.}
\label{fig:prompt_recon_recon}
%
\end{figure*}

\begin{figure*}[!h]
\footnotesize

\begin{tcblisting}{text only,
    halign=left, 
    title=\textbf{GAAR: Prompt for Validity Judgment \& Premise Pruning}, 
    colbacktitle=gray!30!white, 
    coltitle=black,
}
\#\# Defined Variables/Predicates\\
{[[DEFINITION]]}\\
~\\

\#\# Formalized Premises\\
{[[PREMISES]]}\\
~\\

\#\# Formalized Conclusion\\
{[[CONCLUSION]]}\\
~\\

\#\# Deductive Proof\\
{[[PROOF]]}\\
~\\

First, determine necessary formalized premises for the given deductive proof. This includes:\\
~1. Add any missing formalized premises that are necessary to prove the conclusion but cannot be derived from the formalized premises.
~2. Keep all formalized premises that contribute to proving the conclusion through ANY valid reasoning path, even if there are multiple independent paths to the same conclusion. For example, if both ``A, A→C'' and ``B, B→C'' lead to the final conclusion C, keep all premises involved in both paths (A, A→C, B, B→C), not just one path (A, A→C).\\
~3. Remove only those formalized premises that are completely irrelevant and do not contribute to proving the conclusion through any reasoning path.\\
You should format these premises into a python dictionary where keys and values are python strings.\\
~\\

Second, write a python program using z3 that inputs the necessary formalized premises and formalized conclusion and outputs:\\
~1. Their validity, formatted as a python string of either ``valid'' or ``invalid''.\\
~2. All necessary formalized premises that appear in at least one minimal valid reasoning path (i.e., the union of all minimal sets), formatted as a python list of keys of the python dictionary of the necessary formalized premises.\\
You should therefore print two things (a python string and a python list) separately. Please use the below python code snippet.\\
\{Code snippet for validity judgment and premise pruning\}\\
~\\

Third, return the final formalized conclusion that is used in the python program in step 2.\\
~\\

Your response format should be as follows.\\
~\\

\#\#\# Necessary Formalized Premises\\
\verb|```|~python\\
\{\\
~~~~``[Symbol of a premise]'': ``{[Formalization of a premise]}'',\\
~~~~...\\
\}\\
\verb|```|\\
~\\

\#\#\# Python Program\\
\verb|```|~python\\
{[a python program]}\\
\verb|```|\\
~\\

\#\#\# Final Formalized Conclusion\\
{[Formalized conclusion in the python program]}
\end{tcblisting}
\vspace{-2mm}
\caption{Prompt used for validity judgment and premise pruning.}
\label{fig:prompt_recon_validity}
%
\end{figure*}

\begin{figure*}[!h]
\footnotesize

\begin{tcblisting}{text only,
    halign=left, 
    title=\textbf{GAAR: Code Snippet for Validity Judgment \& Premise Pruning}, 
    colbacktitle=gray!30!white, 
    coltitle=black,
}
\verb|```|~python\\
from z3 import *\\
import itertools\\
~\\

\#\#\#\#\#\#\#\#\#\#\#\#\#\#\#\#\#\#\#\#\#\#\#\#\#\#\#\#\#\#\#\#\#\\
\#\#\# Write down your code here \#\#\#\\
\#\#\#\#\#\#\#\#\#\#\#\#\#\#\#\#\#\#\#\#\#\#\#\#\#\#\#\#\#\#\#\#\#\\
~\\

\# Check validity of the argument\\
def check\_validity(premises\_dict, conclusion, solver=None):\\
~~~~if solver is None:\\
~~~~~~~~s = Solver()\\
~~~~else:\\
~~~~~~~~s = solver\\
~~~~~~~~s.reset()\\
~~~~s.add(list(premises\_dict.values()))\\
~~~~s.add(Not(conclusion))\\
~~~~if s.check() == unsat:\\
~~~~~~~~return ``valid''\\
~~~~else:\\
~~~~~~~~return ``invalid''\\
~\\

\# Find all minimal sets and return union of all premises\\
def find\_all\_minimal\_premises(premises\_dict, conclusion):\\
~~~~minimal\_sets = []\\
~~~~premise\_keys = list(premises\_dict.keys())\\
~~~~n = len(premise\_keys)\\
~\\
    
~~~~\# Reuse solver for efficiency\\
~~~~solver = Solver()\\
~\\
    
~~~~\# Check all possible subset sizes from smallest to largest\\
~~~~for subset\_size in range(1, n + 1):\\
~~~~~~~~for subset in itertools.combinations(premise\_keys, subset\_size):\\
~~~~~~~~~~~~subset\_set = set(subset)\\
~\\
            
~~~~~~~~~~~~\# Pruning: Skip if this is a superset of an already found minimal set\\
~~~~~~~~~~~~is\_superset\_of\_minimal = any(minimal\_set.issubset(subset\_set) for minimal\_set in minimal\_sets)\\
~\\
            
~~~~~~~~~~~~if is\_superset\_of\_minimal:\\
~~~~~~~~~~~~~~~~continue\\
~\\
            
~~~~~~~~~~~~subset\_premises = {key: premises\_dict[key] for key in subset}\\
~\\
            
~~~~~~~~~~~~\# Check if this subset is valid\\
~~~~~~~~~~~~if check\_validity(subset\_premises, conclusion, solver) == ``valid'':\\
~~~~~~~~~~~~~~~~minimal\_sets.append(subset\_set)\\
~\\

~~~~\# Return union of all premises in minimal sets\\
~~~~all\_relevant\_premises = set()\\
~~~~for minimal\_set in minimal\_sets:\\
~~~~~~~~all\_relevant\_premises.update(minimal\_set)\\
~\\
    
~~~~return sorted(list(all\_relevant\_premises))\\
~\\

validity = check\_validity(premises, conclusion)\\
print(validity)\\
minimal\_premises = find\_all\_minimal\_premises(premises, conclusion)\\
print(minimal\_premises)\\
\verb|```|~\\
~\\
\end{tcblisting}
\vspace{-2mm}
\caption{Python code snippet for validity judgment and premise pruning.}
\label{fig:prompt_recon_validity_code}
%
\end{figure*}

\begin{figure*}[!h]
\footnotesize

\begin{tcblisting}{text only,
    halign=left, 
    title=\textbf{GAAR: Prompt for Streamlining}, 
    colbacktitle=gray!30!white, 
    coltitle=black,
}
\#\# Defined Variables/Predicates\\
{[[DEFINITION]]}\\
~\\

\#\# Formalized Premises\\
{[[PREMISES]]}\\
~\\

\#\# Formalized Conclusion\\
{[[CONCLUSION]]}\\
~\\

Given definitions of variables and/or predicates, generate natural language (NL) descriptions of formalized premises and conclusion. Your response format should be as follows. Provide only the natural language descriptions without including any propositional variables, logical symbols, or formulas, unless absolutely necessary.\\
~\\

\#\#\# NL Premises\\
{[list of premises in natural language]}\\
~\\

\#\#\# NL Conclusion\\
{[conclusion in natural language]}
\end{tcblisting}
\vspace{-2mm}
\caption{Prompt used for streamlining.}
\label{fig:prompt_recon_deformalization}
%
\end{figure*}

\begin{figure*}[!h]
\footnotesize

\begin{tcblisting}{text only,
    halign=left, 
    title=\textbf{GAAR: Prompt for Faithfulness Judgment}, 
    colbacktitle=gray!30!white, 
    coltitle=black,
}
\# Debate Topic\\
{[[TOPIC]]}\\
~\\

% \# Background\\
% {[[BACKGROUND]]}\\
% ~\\

\# Argument\\
{[[ARGUMENT]]}\\
~\\

\# Argument Reconstruction\\
\#\# Premises\\
{[[PREMISES]]}\\
~\\

\#\# Conclusion\\
{[[CONCLUSION]]}\\
~\\

For an argument from the debate, its reconstruction as a premise-conclusion structure is given. Note that the argument is {[[ARG\_TYPE]]}. Your task is to judge whether the construction is faithful or not. You should judge the faithfulness according to the following three criteria:\\
~- \textbf{**Accuracy \& Charity.**} The reconstruction should keep the author’s intended meaning while eliminating irrelevancies—i.e., obey the principle of charity and prefer the strongest sensible reading of ambiguous passages.\\
~~~- If a premise expresses or entails a universal quantifier, determine whether the universal quantifier was used as intended in the original argument, or whether the universal quantifier over-generalized the premise beyond the original argument's intent.\\
~~~~~- Definition of ``universal quantifier'': A universal quantifier is a logical operator that expresses a predicate is true for every member of a universe of discourse. It is symbolized as $\forall$, or expressed through words like `for all x', `every', `any', etc. This should be distinguished from statements that are merely ``universal'' in nature; that is, general or broadly applicable statements that lack explicit universal quantifiers.\\
~~~~~- First, determine if a universal quantifier is actually present by definition.\\
~~~~~- When evaluating over-generalization: If a universal quantifier is used in a way that over-generalizes **beyond the original argument's intent**, then the premise must be made more specific. For example, suppose the reconstructed premises are ``Person P has property A.'', ``For all x, if x has property A, then x has property B.'', and ``Therefore, Person P has property B.'' However, the original argument may only intend to claim that Person P having property A leads to Person P having property B. So, the reconstruction is not faithful and should be revised to be specific: ``If Person P has property A, then Person P has property B.'', without the universal quantifier.\\
~- \textbf{**Completeness.**} All explicit premises, the main conclusion, and any indispensable implicit premises must be included.\\
~- \textbf{**Parsimony.**} The reconstruction must avoid including unnecessary premises.\\
~~~- Premises that introduce content not present in the original argument or that serve as supporting examples (rather than independent premises) are unnecessary and should be excluded, but bridging premises needed to make the logical structure explicit are necessary and should be retained.\\
~\\

Do not include logical validity while judging the faithfulness, especially if the argument is formally fallacious.\\
~\\

The output format should be as follows.\\
~\\

\# Reasoning\\
{[Explain step-by-step]}\\
~\\

\# Faithfulness\\
{[Yes or No]}
\end{tcblisting}
\vspace{-2mm}
\caption{Prompt used for faithfulness judgment.}
\label{fig:prompt_recon_faithfulness}
%
\end{figure*}

\begin{figure*}[!h]
\footnotesize

\begin{tcblisting}{text only,
    halign=left, 
    title=\textbf{LLM Prompting: Prompt for Argument Reconstruction}, 
    colbacktitle=gray!30!white, 
    coltitle=black,
}
\# Debate Topic\\
{[[TOPIC]]}\\
~\\

% \# Background\\
% {[[BACKGROUND]]}\\
% ~\\

\# Argument\\
{[[ARGUMENT]]}\\
~\\

You are given a debate topic and an argument from that debate topic. First, identify what type(s) of reasoning this argument uses from the four types listed below. Note that the argument may contain multiple types of reasoning. Then reconstruct the argument with a premise-conclusion structure. If the debate topic has clear pro/con positions, determine the conclusion by taking either a positive or negative stance. Add implicit premises and intermediate conclusions if needed.\\
~\\

\#\# Argument Types\\
\{Description of 4 types from the \textit{general} theory or 60 types from the \textit{specific} theory\}\\
~\\

\#\# Reconstruction Guidelines Based on Argument Types\\
\{Reconstructions of each argument type\}\\
~\\

Your output should be composed of the argument reconstruction. List premises and a conclusion.\\
The output format should be as follows.\\
~\\

\# Argument Reconstruction\\
~\\

\#\# Premises\\
{[list of explicit and implicit premises]}\\
~\\

\#\# Conclusion\\
{[a conclusion]}
\end{tcblisting}
\vspace{-2mm}
\caption{Prompt used for reconstructing an argument (LLM Prompting).}
\label{fig:prompt_recon_recon_llmprompting}
%
\end{figure*}

\begin{figure*}[!h]
\footnotesize

\begin{tcblisting}{text only,
    halign=left, 
    title=\textbf{Evaluation: Prompt for Validity Judgment (Only for LLM Prompting)},
    colbacktitle=gray!30!white, 
    coltitle=black,
}
\# Debate Topic\\
{[[TOPIC]]}\\
~\\

% \# Background\\
% {[[BACKGROUND]]}\\
% ~\\

\# Argument\\
{[[ARGUMENT]]}\\
~\\

\# Argument Reconstruction\\
{[[ARGUMENT\_RECONSTRUCTION]]}\\
~\\

You are given a debate topic and an argument from that debate topic, and the argument reconstruction with a premise-conclusion structure.\\
~\\

First, define variables and/or predicates and formalize premises and the conclusion in argument reconstruction.\\
~\\

IMPORTANT: The argument reconstruction may not be valid. STILL you must formalize it *faithfully as-is*, remaining invalid in the same way.\\
~- Do NOT change the meaning of any premise or the conclusion.\\
~- Do NOT add, remove, or reorder premises.\\
~- Do NOT merge, split, or rewrite premises.\\
~- Each natural-language premise must map to exactly one formalized premise.\\
~\\

Second, by utilizing the definition and formalized premises/conclusion in the first step, write a python program using z3 that inputs the formalized premises and formalized conclusion and outputs:\\
~1. Their validity, formatted as a python string of either ``valid'' or ``invalid''.\\
~2. All necessary formalized premises that appear in at least one minimal valid reasoning path (i.e., the union of all minimal sets), formatted as a python list of keys of the python dictionary of the necessary formalized premises.\\


You should therefore print a python string. Please use the below python code snippet.\\
\{Code snippet for validity judgment and premise pruning\}\\
~\\

Your response format should be as follows.\\
~\\

\#\# Defined Variables/Predicates\\
{[[DEFINITION]]}\\
~\\

\#\# Formalized Premises\\
{[[PREMISES]]}\\
~\\

\#\# Formalized Conclusion\\
{[[CONCLUSION]]}\\
~\\

\#\#\# Python Program\\
\verb|```|~python\\
{[a python program]}\\
\verb|```|~\\
\end{tcblisting}
\vspace{-2mm}
\caption{Prompt used for evaluating validity (LLM Prompting).}
\label{fig:prompt_eval_validity}
%
\end{figure*}

\begin{figure*}[!h]
\footnotesize

\begin{tcblisting}{text only,
    halign=left, 
    title=\textbf{Evaluation: Prompt for Pairwise Faithfulness Judgment (1/2)},
    colbacktitle=gray!30!white, 
    coltitle=black,
}
You are an expert in argument analysis and logical reasoning. Your task is to *critically* examine two proposed argument reconstructions of the same source argument, and decide on which is more faithful to the original argument (and by how much) on several criteria.\\
~\\

Your task is to assess which reconstruction more faithfully captures what the original argument ACTUALLY SAYS AND DOES—not which reconstruction presents a stronger, more valid, or more charitable version of the argument. A faithful reconstruction captures the argument as it actually appears in the source, including any logical gaps, non-sequiturs, or fallacious reasoning patterns.\\
~\\

\# Debate Topic\\
{[[TOPIC]]}\\
~\\

% \# Background\\
% {[[BACKGROUND]]}\\
% ~\\

\# Argument\\
{[[ARGUMENT]]}\\
~\\

\# Reconstruction A\\
{[[RECONSTRUCTION\_A]}\\
~\\

\# Reconstruction B\\
{[[RECONSTRUCTION\_B]}\\
~\\

Compare the relative quality of each reconstruction on the following criteria:\\
~\\

1. Accuracy\\
- Assess whether the reconstruction accurately represents the original argument's actual reasoning path, including any inferential leaps, gaps, or logical fallacies, without misrepresentation.\\
- Misrepresentation includes both distorting what was said AND artificially strengthening weak or fallacious reasoning.\\
- Do NOT reward a reconstruction for ``fixing'' or ``improving'' the original argument's logic.\\
- If both reconstructions preserve the meaning and actual logical structure of the original argument without misrepresentation, the result should be a tie on accuracy.\\
~\\

2. Completeness\\
- Assess whether all essential or core premises required to reconstruct the original argument are included.\\
- If the original argument has logical gaps, a complete reconstruction captures those gaps rather than filling them.\\
- If both reconstructions include all essential or core premises required to reconstruct the original argument, the result should be a tie on completeness.\\
~\\

3. Parsimony\\
- Assess whether the reconstruction avoids including premises that are unnecessary for representing the original argument's actual reasoning.\\
~~- Do NOT judge the reconstruction as more parsimonious simply because it has less number of premises. As long as premises are necessary, the number of premises does not matter.\\
~~- Premises that introduce content or claims not present (either explicitly or implicitly) in the original argument are unnecessary.\\
~~- Premises that appear in the original text but serve merely as supporting details, examples, or evidence that strengthen the plausibility of a stated premise—rather than functioning as independent premises that directly support the conclusion—are unnecessary.\\
~~- Premises added to ``fix'' logical gaps or fallacies that exist in the original argument are unnecessary.\
~~- Bridging premises are acceptable ONLY if they make explicit what is clearly implicit in the original argument, as long as they preserve the logical structure of the original argument. Bridging premises that would repair a fallacy or fill a logical gap that the original argument does not address are unnecessary additions and should be penalized.\\
- If both reconstructions avoid unnecessary premises, the result should be a tie on parsimony.\\
\end{tcblisting}
\vspace{-2mm}
\caption{Prompt used for pairwise faithfulness judgment (1/2).}
\label{fig:prompt_eval_faithfulness_1}
%
\end{figure*}

\begin{figure*}[!h]
\footnotesize

\begin{tcblisting}{text only,
    halign=left, 
    title=\textbf{Evaluation: Prompt for Pairwise Faithfulness Judgment (2/2)},
    colbacktitle=gray!30!white, 
    coltitle=black,
}
Judging notes:\\
- Do NOT assess logical validity, soundness, or persuasiveness as virtues of a reconstruction.\\
- ONLY assess faithfulness to what the original argument says and how it is actually structured.\\
- Do NOT penalize a reconstruction for containing fallacies if that fallacy exists in the original argument.\\
- Do NOT reward a reconstruction for eliminating fallacies that exist in the original argument.\\
- Length by itself is irrelevant to all three criteria. Do not reward or penalize a reconstruction for being longer or shorter unless length directly results in misrepresentation, omission of essential premises, or inclusion of unnecessary premises.\\
- The three criteria (accuracy, completeness, parsimony) are independent, i.e., a reconstruction may perform well on one and poorly on another.\\
~\\

For each criterion, output either:\\
- the winning reconstruction (A or B) followed by a + / ++ / +++ / ++++ / +++++ disparity rating, or\\
- ``TIE'' if neither reconstruction is more faithful on that criterion.\\
~\\

The overall\_winner should reflect which reconstruction is more faithful in aggregate across the three criteria, taking into account both (i) which criteria each reconstruction wins and (ii) the relative magnitude of the disparity ratings (+ to +++++) on those criteria.\\
~\\

A reconstruction with fewer criterion wins may still be selected as the overall\_winner if its advantage on one or more criteria is substantially stronger. In such cases, the reasoning must explicitly justify why that criterion (or those criteria) should be weighted more heavily for faithfulness to the original argument in this specific comparison.\\
~\\

If, after considering both the number of criteria won and the relative strength of those wins, neither reconstruction is clearly more faithful overall, output ``TIE''.\\
~\\

Examples:\\
``accuracy'': ``A++''\\
``completeness'': ``TIE''\\
``parsimony'': ``B++++''\\
``overall\_winner'': ``B''\\
~\\

Respond in valid json without additional commentary (remembering to escape any string quotes), in this format:\\
~\\

\{\\
~~``reasoning": ``Detailed reasoning comparing the two reconstructions across accuracy, completeness, and parsimony.",\\
~~``accuracy": ``winner \& disparity rating OR TIE",\\
~~``completeness": ``winner \& disparity rating OR TIE",\\
~~``parsimony": ``winner \& disparity rating OR TIE",\\
~~``overall\_winner": ``winner OR TIE''\\
\}
\end{tcblisting}
\vspace{-2mm}
\caption{Prompt used for pairwise faithfulness judgment (2/2).}
\label{fig:prompt_eval_faithfulness_2}
%
\end{figure*}

\end{document}